\documentclass[final,3p,times,twocolumn,authoryear]{elsarticle}

\usepackage{bm}
\usepackage{newfloat}
\usepackage{amssymb}
\usepackage{amsmath}
\usepackage{bigints}
\usepackage{tikz-qtree}
\usepackage[linesnumbered]{algorithm2e}
\usepackage{graphicx}
\usepackage{subfig}
\usepackage[font={small,sl}]{caption}
\usepackage{epstopdf}
\usepackage{booktabs}
\usepackage{array}
\usepackage{booktabs}
\usepackage{longtable}
\usepackage[tc]{titlepic}
\usepackage{algorithm2e}
\usepackage{graphicx}
\usepackage{soul}
\usepackage{xcolor}
\usepackage{colortbl}
\usepackage[titletoc,toc,title]{appendix}
\usepackage[subfigure]{tocloft}
\usepackage{natbib}
\usepackage{multirow}
\usepackage{cleveref}
\usepackage{subfig}	

\newcommand*\xbar[1]{
   \hbox{
     \vbox{
       \hrule height 0.5pt 
       \kern0.5ex             
       \hbox{
         \kern-0.26em         
         \ensuremath{#1}
         \kern-0.26em        
       }
     }
   }
} 
	
\begin{document}

\begin{frontmatter}

\title{Generative diffeomorphic atlas construction from brain and spinal cord MRI data}

\author[label1]{Claudia Blaiotta\corref{cor1}}
\ead{claudia.blaiotta.13@ucl.ac.uk}
\cortext[cor1]{Corresponding author}

\author[label1,label2]{Patrick Freund}

\author[label3]{M. Jorge Cardoso}

\author[label1]{John Ashburner}

\address[label1]{Wellcome Trust Centre for Neuroimaging, University College London, London, UK}
\address[label2]{Spinal Cord Injury Center Balgrist, University Hospital Zurich, University of Zurich, Zurich, Switzerland}
\address[label3]{Translational Imaging Group, CMIC, University College London, London, UK}

\begin{abstract}
    In this paper we will focus on the potential and on the challenges associated with the development of an integrated brain and spinal cord modelling framework for processing MR neuroimaging data. 
	The aim of the work is to explore how a hierarchical generative model of imaging data, which captures simultaneously the distribution of signal intensities and the variability of anatomical shapes across a large population of subjects, can serve to quantitatively investigate, \textit{in vivo}, the morphology of the central nervous system (CNS). 
	In fact, the generality of the proposed Bayesian approach, which extends the hierarchical structure of the segmentation method implemented in the SPM software, allows processing simultaneously information relative to different compartments of the CNS, namely the brain and the spinal cord, without having to resort to organ specific solutions (e.g. tools optimised only for the brain, or only for the spinal cord), which are inevitably harder to integrate and generalise.	
\end{abstract}


\end{frontmatter}

	\section{Introduction}
	
	The spinal cord is a long and thin cylindrical structure of the central nervous system, which constitutes the main pathway for transmitting information between the brain and the rest of the body. Not only is the spinal cord a major site of traumatic injury (SCI), but it can also be affected by a number of neurodegenerative diseases, such as multiple sclerosis, amyotrophic lateral sclerosis, transverse myelitis and neuromyelitis optica \citep{rocca2015spinal}. Indeed, the spinal cord is a clinically eloquent structure, since trauma, ischemia and inflammation can affect the cord at any level, thus resulting in impairment of motor, sensory and autonomic functions \citep{huber2015tracking,freund2013tracking}.
	
	Understanding these degenerative processes represents a crucial step towards the development of effective therapeutic interventions, as well as towards the identification of sensitive and selective diagnostic criteria. In particular, quantification of spinal cord tissue loss (i.e. atrophy) has been regarded over the past two decades as a promising  biomarker, which could potentially help in monitoring disease progression, predicting clinical outcome and understanding the mechanisms underlying neurological disability (e.g. demyelination, inflammation, axonal or neuronal loss), in a number of conditions that affect the central nervous system both at the brain and spinal cord level, such as multiple sclerosis (MS) and traumatic spinal cord injury (SCI) \citep{miller2002measurement,freund2013mri,freund2013tracking,grossman2000assessment,bakshi2005measurement}. 
	
    Since the spinal canal is surrounded and protected by a thick vertebral bone layer, neuroimaging techniques, particularly MRI, represent the most effective tools to investigate non-invasively and \textit{in vivo} the structure and function of the spinal cord, both in physiological and pathological conditions.	Unfortunately spinal cord MRI is not immune from technical challenges. Some of them are intrinsic to MR imaging, such as the presence of intensity inhomogeneities, while others arise from the peculiar anatomy of the cord itself, for instance from its small cross-sectional area \citep{grossman2000assessment,wheeler2014current,stroman2014current}.	Nevertheless, spinal cord imaging using MR techniques has improved significantly over the past few years, especially with the introduction of phased-array surface coils and fast spin-echo sequences \citep{stroman2014current}. 
	
	Further advances in the field of spinal cord MRI are encouraged by the fact that significant correlations between spinal cord atrophy measures, obtained from imaging data, and indicators of neurological impairment, such as motor or sensory function scores, have been shown and reproduced, within multiple spinal cord imaging studies \citep{losseff1998measures,kidd1993spinal,filippi1996spinal,losseff1996spinal,freund2013mri,grabher2015tracking}.
	
	Within these type of studies, delineating the cord represents the first step for assessing atrophy or detecting any other morphometric change, or difference. This indicates that there is an urgent need not only for automated algorithmic solutions dedicated to spinal cord tissue classification and image registration \citep{chen2013automatic,van2005semi,fonov2014framework,levy2015white,taso2014construction,de2017sct}, but also for large, systematic and reproducible validation studies to objectively assess the performance of such tools \citep{scchallenge}.

	Not surprisingly, the first methods that appeared in the literature to perform spinal cord image segmentation and the subsequent  volumetric analyses were based on semi-automated algorithms. Among these, one of the earliest is described in the work of \citet{coulon2002quantification}, where they introduce an algorithm for fitting a cylindrical cubic B-spline surface to MR spinal cord images.
	Later on, a few other semi-automated solutions have been presented by \citet{van2005semi} and \cite{horsfield2010rapid}. All of these methods require that the user approximately marks the cord centre, so as to provide a reliable initialisation of the algorithms.

	Only very recently have fully automated spinal cord segmentation methods started to be proposed.	
	\cite{chen2013automatic} introduced a fuzzy c-means algorithm with topological constraints to segment the cervical and thoracic spinal cord from MR images. Their method relies on a statistical atlas of the cord and the surrounding CSF, which is constructed from only five manual segmentations.
	Instead, \cite{de2014robust} proposed a fully automated method for delineating the contour of the spinal cord, in T1- and T2-weighted MR images, by warping of a deformable cylindrical model.
	
	The first significant effort to define and introduce a standard anatomical space for spinal cord neuroimaging studies relates to the work of \cite{fonov2014framework}, who developed a standard stereotactic space for spinal cord imaging data, between the vertebral levels of C1 and T6 (MNI-Poly-AMU template). 
	
	Their template is generated using the image registration algorithm presented in \cite{avants2008symmetric} and includes a T2-weighted average image, together with probabilistic gray and white matter maps. Such tissue probability maps were developed by \cite{taso2014construction}, via automated registration of manually labelled MRI scans of 15 subjects.
	
	The work of \cite{fonov2014framework,levy2015white,taso2014construction} constitutes an important step towards the development of robust and reliable tools for analyzing structural spinal cord data. Indeed, having a common anatomical framework can potentially allow the comparison of results obtained by different research groups on different data sets, thus speeding up the progress of spinal cord imaging research. 
	
	The work presented here aims to provide researchers with a general and comprehensive modelling framework to interpret large data sets of MRI scans from a Bayesian generative perspective. This is achieved by building on the modelling elements introduced in \citet{unified,ashburner2011diffeomorphic,blaiotta2016variational}, which are further expanded here and integrated in one single algorithmic framework. 
	
	The aim is to demonstrate the validity of such a generative approach, especially for the purpose of performing simultaneous brain and spinal cord morphometric analyses using MRI data sets. 
	In doing so, a strategy is outlined on how to overcome some of the limitations of most currently available image processing tools for neuroimaging, whose performance has been optimised on the brain at the expense of the spinal cord (indeed the spinal cord is frequently neglected \textit{tout court} by such tools).
	
	\section{Methods}
	\label{methods}
	
	Let us consider a population of $M$ subjects belonging to a homogeneous group, from an anatomical point of view, and let us assume that $D$ image volumes of different contrast are available for each subject. 
	
	From a generative perspective, the image intensities $\xbar{\mathbf{{X}}} = \{\mathbf{
	X}_i\}_{i=1,\ldots,M}$, which constitute the observed data, can be thought of as being generated by sampling from $D$-dimensional Gaussian mixture probability distributions, after non-linear warping of a probabilistic anatomical atlas \citep{evans1994mri}. 
	
	Such an atlas carries \textit{a priori} anatomical knowledge, in the form of average shaped tissue probability maps. From a mathematical modelling point of view, the atlas encodes local (i.e. spatially varying) mixing proportions $\Theta_\pi =\{\bm{\pi}_j\}_{j=1,\ldots,N_\pi}$ of the mixture model, with $j$ being an index set over the $N_\pi$ template voxels, as detailed in the following subsection.
	
	\subsection{Tissue priors}
	\label{prior}
	
	Each image voxel $j \in \{1,\ldots,N_i\}$, for each subject $i \in \{1,\ldots,M\}$ is considered as being drawn from $K$ possible tissue classes. The following prior latent variable model defines the probability of finding tissue type $k$, at a specific location $j$ (i.e. centre of voxel $j$), in image $i$, prior to observing the corresponding image intensity signal
	 
	\begin{align}
	p({z}_{ijk}=1|\Theta_\pi,\Theta_w, \Theta_u) = \frac{w_{ik}\,\pi_{k}(\bm{\xi}_{i}(\mathbf{y}_j))}{\sum_{c=1}^K w_{ic}\,\pi_{c}(\bm{\xi}_{i}(\mathbf{y}_j))}\;,
	\label{eq:prz}
	\end{align}
	or equivalently
	\begin{align}
	p(\mathbf{{z}}_{ij}|\Theta_\pi,\Theta_w,\Theta_u) = \prod_{k=1}^K \left(\frac{w_{ik}\,\pi_{k}(\bm{\xi}_{i}(\mathbf{y}_j))}{\sum_{c=1}^K w_{ic}\,\pi_{c}(\bm{\xi}_{i}(\mathbf{y}_j))}\right)^{z_{ijk}}\;.
	\label{eq:mul}
	\end{align}
	
	Class memberships, for each subject and each voxel, are encoded in the latent variable $\mathbf{z}_{ij}$, which is a $K$-dimensional binary vector.
	$\{\pi_k\}_{k=1,\ldots,K}$ are scalar functions of space $\pi_k:\Omega_\pi \rightarrow \mathbb{R}$, common across the entire population, which satisfy the constraint
	\begin{equation}
	\sum_{k=1}^K \pi_k(\bm{y}) =1 \;,~\forall \bm{y} \in \Omega_\pi \subset \mathbb{R}^3\;,
	\label{eq:pin}
	\end{equation}
	with $\bm{y}$ being a continuous coordinate vector field. Global weights $\Theta_w=\{\bm{w}_i\}_{i=1,\ldots,M}$ are introduced to further compensate for individual differences in tissue composition.
	
	 In equation \eqref{eq:prz}, $\bm{\xi}_{i}$ denotes a generic spatial transformation, parametrised by $\Theta_u$, which allows projecting prior anatomical information onto individual data, with $\bm{\xi}_{i}: \Omega_i \rightarrow \Omega_\pi$ being a continuous mapping from the domain $\Omega_i \subset \mathbb{R}^3$ of image $i$, into the space of the tissue priors $\Omega_\pi \subset \mathbb{R}^3$. 
	
	Since digital image data is a discrete signal, defined on a tridimensional voxel grid, each mapping $\bm{\xi}_i$ needs to be discretised as well, via sampling at the centre of every voxel $j \in\{1,\ldots,N_i\}$, to give the discrete mapping $\{\bm{\xi}_{i}(\mathbf{y}_j)\}_{j=1,\ldots,N}$ that appears in \eqref{eq:prz}.
	
	As opposed to the modelling approach described in \citet{unified}, where the tissue priors were considered as fixed and known \textit{a priori} quantities, here the tissue probability maps are treated as random variables, whose point estimates or full posteriors can be inferred via model fitting \citep{bhatia2007groupwise,ribbens2014unsupervised}. 
	
	For this purpose, a finite dimensional parametrisation of $\{\pi_k\}_{k=1,\ldots,K}$ needs to be defined. Typically, whenever a continuous function needs to be reconstructed from a finite discrete sequence, it is possible to formulate the problem as an interpolation that makes use of a finite set of coefficients and continuous basis functions. Since the priors $\{\pi_k\}_{k=1,\ldots,K}$ are bounded to take values in the interval $[0,1]$ on the entire domain $\Omega_\pi$ (see equation \eqref{eq:pin}), not all basis functions are well suited here. Linear basis functions, besides being quite a computationally efficient choice, have the convenient property of preserving the values of $\{\pi_k\}_{k=1,\ldots,K}$ in the interval $[0,1]$, as long as the coefficients are also in the same interval. Such coefficients belong to the discrete set $\Theta_\pi=\{\bm{\pi}_{j}\}_{j=1,\ldots,N_\pi}$ of $K$-dimensional vectors, with
	\begin{align}
	\sum_{k=1}^{K} \pi_{jk} = 1,~\;\forall j \in \{1,\ldots,N_\pi\}\;.
	\end{align}
	 They can be learned directly from the data, as it will be shown in the following section.
	
	Additionally, prior distributions on the parameters $\{\bm{\pi}_{j}\}_{j=1,\ldots,N}$ can be introduced \citep{bishop2006pattern}. Dirichlet priors are the most convenient choice here, since they are conjugate to multinomial forms of the type in \eqref{eq:mul}, and they can be expressed as 
	\begin{equation}
	p(\bm{\pi}_j) = \text{Dir}(\bm{\pi}_j|\bm{\alpha}_0) = C(\bm{\alpha}_0) \prod_{k=1}^K \pi_{jk}^{\alpha_k-1}\;,
	\end{equation}  
	where the normalising constant is given by
	\begin{equation}
	C(\bm{\alpha}_0) = \frac{\Gamma(\bar \alpha)}{\Gamma(\alpha_1) \ldots \Gamma(\alpha_k)}\;,
	\end{equation}
	with $\Gamma(\cdot)$ being the gamma function and
	\begin{equation}
	\bar \alpha = \sum_{k=1}^K \alpha_k\;.
	\end{equation}
	
	\subsection{Diffeomorphic image registration}
	
	As anticipated in the previous sections, the generative interpretation of imaging data that this work relies on involves warping an unknown, average-shaped atlas to match a series of individual scans. 

	Such a problem, that is to say template matching via non-rigid registration, has been largely explored in medical imaging, mainly for solving image segmentation or structural labeling problems, in an automated fashion \citep{unified,shen2004measuring,christensen1999consistent,chui2001unified,bajcsy1983computerised,iglesias2012generative,pluta2009appearance,warfield1999nonlinear,khan2008freesurfer,bowden1998anatomical}.
	
	Indeed, the modelling of spatial mappings between different anatomies can be approached in a variety of manners, depending on the adopted model of shape and on the objective function (i.e. similarity metric and regularisation) that the optimisation is based on, thus leading to a variety of algorithms with remarkably different properties \citep{penney1998comparison,denton1999comparison,klein2009evaluation}.

	The work presented here is formulated according to the Large Deformation Diffeomorphic Metric Mapping (LDDMM) framework \citep{younes2010shapes}, where the transformations mapping between the source images and the target image are assumed to belong to a Riemannian manifold \footnote{A Riemannian manifold, in differential geometry, is a smooth manifold $M$ equipped with a Riemannian metric (inner product). In particular, the Riemannian metric $G_p$ on the $n$-dimensional manifold $M^n$ defines, for every point $p \in M$, the scalar product of vectors in the tangent
    space $T_pM$, in such a way that given two vectors ${\bm{x},\bm{y}} \in M$, the inner product $G_p(\bm{x},\bm{y})$ depends smoothly on the point $p$. The tangent space represents the nearest approximation of the manifold by a vector space \citep{warner2013foundations}.} of diffeomorphisms \citep{ashburner2007fast}. A diffeomorphism  $\bm{\phi}: \Omega \rightarrow \Omega$ is a smooth differentiable map (with a smooth differentiable inverse $\bm{\phi}^{-1}$) defined on a compact, simply connected domain $\Omega \subset\mathbb{R}^3$. 
	
	One way of constructing transformations belonging to the diffeomorphic group $\text{Diff}(\Omega)$ is to solve the following non-stationary transport equation \citep{joshi2000landmark}
	\begin{equation}
	\frac{\text{d}}{\text{d}t} \bm{\phi}(\bm{y},t) = \bm{u}(\bm{\phi}(\bm{y},t),t),\;\bm{\phi}(\bm{y},0) = \bm{y}, \;t \in [0, 1]\;,
	\label{ode}
	\end{equation}
	where $ \bm{u}(\bm{\phi}(\bm{y},t),t) \in \mathcal{H}$ is a time dependent, smooth velocity vector field, in the Hilbert space\footnote{A Hilbert space $\mathcal{H}$ is a complete inner product space, where an inner product is a map $\langle\cdot, \cdot \rangle : \mathcal{H} \times \mathcal{H} \rightarrow \mathbb{C}$\;, which associates each pair of vectors in the space with a scalar quantity. In particular given $\bm{x}, \bm{y}, \bm{z} \in \mathcal{H}$ and $a,b \in \mathbb{C}$
	\begin{align}
	\langle a\bm{x} + b\bm{y}, \bm{z} \rangle &= a\langle \bm{x}, \bm{z} \rangle + b \langle \bm{y}, \bm{z} \rangle\;,\\
	\langle \bm{x},\bm{x}\rangle &\geq 0,\;\text{and} \;\langle \bm{x},\bm{x}\rangle = 0  \Leftrightarrow \bm{x} = 0\;,\\
	\langle \bm{x},\bm{y} \rangle &= \langle \bm{y},\bm{x} \rangle\;.
	\end{align}
	
    An inner product naturally induces a norm by $||\bm{x}|| = \langle \bm{x}, \bm{x}\rangle^{1/2}$, therefore every inner product space is also a normed vector space \citep{dieudonne2013foundations}.} $\mathcal{H}$.
	
	The initial map, at $t = 0$, is equal to the identity transform $\bm{\phi}(\bm{y},0) = \bm{y}$, while the final map, endpoint of the flow of the velocity field $\bm{u}$, can be computed by integration on the unitary time interval $t \in [0, 1]$ \citep{beg2005computing}.
	
	\begin{equation}
	\bm{\phi}(\bm{y},1) = \int_0^1 \bm{u}(\bm{\phi}(\bm{y},t),t) \text{d}t + 	\bm{\phi}(\bm{y},0)\;.
	\end{equation}
	
	Following from the theorems of existence and uniqueness of the solution of partial differential equations (\textit{p.d.e.}), the solution of \eqref{ode}
	is uniquely determined by the velocity field $\bm{u}(\bm{\phi}(\bm{y},t),t)$  and by the initial condition $\bm{\phi}(\bm{y},0)$.
	
	A diffeomorphic path $\bm{\phi}$ is not only differentiable, but also guaranteed to be a one-to-one mapping. Such a quality is highly desirable for finding morphological and functional correspondences between different anatomies without introducing tears or foldings, which would violate the conditions for topology preservation \citep{christensen1999consistent}. Additionally, the diffeomorphic framework provides metrics to quantitatively evaluate distances between anatomies or shapes.
	It should also be noted that diffeomorphisms are locally analogous to affine transformations \citep{avants2006lagrangian}.
	
	In practice, finding an optimal diffeomorphic transformation to align a pair, or a group, of images involves optimising an objective function (e.g. minimising a cost function), in the space $\mathcal{H}$ of smooth velocity vector fields defined on the domain $\Omega$. The required smoothness is enforced by constructing the norm on the space $\mathcal{H}$ through a differential operator $\mathbf{L_u}$ \citep{beg2005computing}, such that a quantitative measure of smoothness can be obtained via
	\begin{align}
	\mathcal{R}(\mathbf{u}) = ||\mathbf{L_u u}||_{L^2}^2\;,
	\end{align}
	where $\mathbf{u}$ is a discretised version of $\bm{u}$.
	
	The form of the cost function will depend on how the observed data is modelled. For the work presented here, groupwise alignment is achieved via maximisation of the following variational objective function
	\begin{align}
	\begin{split}
	\mathcal{E}(\Theta_u) = & \mathbb{E}_{\mathbf{Z}}[\log p(\xbar{\mathbf{{Z}}}|\Theta_{\pi},\Theta_w,\Theta_{u})]  + \log p(\Theta_u) + \text{const}\\
	= &\sum_{i=1}^M\sum_{j=1}^{N_i} \sum_{k=1}^K \gamma_{ijk}\log\left (\frac{w_{ik}\pi_{k} (\bm{\phi}_{i}(\mathbf{y}_j))}{\sum_{c=1}^K w_{ic}\,\pi_{c}(\bm{\phi}_{i}(\mathbf{y}_j))}\right)\\ - &\frac{1}{2}\sum_{i=1}^{M}||\mathbf{L_{u}}\mathbf{u}_i ||_{L^2}^2 + \text{const}\;,
	\label{eq:of}
	\end{split}
	\end{align}
	where $\xbar{\mathbf{{Z}}} = \{\mathbf{
	Z}_i\}_{i=1,\ldots,M}$ is the set of latent variables across the entire population, $\{\bm{\gamma}_{ij}\}_{i,j} = \left\{\mathbb{E}[\mathbf{z}_{ij}]\right\}_{i,j}$ are $K$-dimensional vectors of posterior belonging probabilities, $\Theta_\pi$ indicates the coefficients used to parametrise the tissue priors $\{\pi_k\}_{k=1,\ldots,K}$ and $\Theta_w$ denotes a set of individual tissue weights $\{\bm{w}_i\}_{i=1,\ldots,M}$ for rescaling the tissue probability maps.
	The coordinate mappings $\{\bm{\phi}_i\}_{i=1,\ldots,M}$ are encoded in the parameter set $\Theta_u$, which consists of $M$ vectors of coefficients $\{\mathbf{u}_{i}\}_{i=1,\ldots,M}$, containing $3\times N_i$ elements  each. Such coefficients can be used to construct continuous initial velocity fields via trilinear, or higher order, interpolation.
	
	A procedure known as geodesic shooting \citep{miller2006geodesic,ashburner2011diffeomorphic,allassonniere2005geodesic,vialard2012diffeomorphic,beg2006computing} is applied, within the work presented here, to compute diffeomorphic deformation fields from corresponding initial velocity fields. Such a procedures exploits the principle of conservation of momentum \citep{younes2009evolutions}, which is given by $\mathbf{m}_t = \mathbf{L_u^\dagger L_u u}_t$, with $\mathbf{L_u^\dagger}$ being the adjoint of the differential operator $\mathbf{L_u}$, to integrate the dynamical system governed by \eqref{ode} without having to store an entire time series of velocity fields.
	The implementation adopted here relies on the work presented in \cite{ashburner2011diffeomorphic}.
	
	The posterior membership probabilities $\{\bm{\gamma}_{ij}\}_{i,j}$ that appear in \eqref{eq:of} can be computed by combining the prior latent variable model introduced in \ref{prior} with a likelihood model of image intensities, which will be described in sub\cref{secIn}, thus leading to a fully unsupervised learning scheme. 
	
	Alternatively, when manual labels are available, binary posterior class probabilities can be derived directly from such categorical annotations, without performing inference from the observed image intensity data. In particular, if all input data has been manually labelled, then the resulting algorithm would implement a fully supervised learning strategy, while, if only some of the data has associated training labels, a hybrid approach can be adopted, which would fall into the category of semisupervised learning \citep{chapelle2006semi,filipovych2011semi}.
	
	Finally, it is also possible to take into account the uncertainty inherent in the process of manual rating. In such a case, the actual posterior probabilities can be computed by making use of the categorical output of manual labelling together with an estimate of the rater sensitivity and with a generative intensity model.
	
	Making use of Bayes rule, this gives
	\begin{align}
	\begin{split}
	\gamma_{ijk} &= p (z_{ijk}=1|\mathbf{x}_{ij},\Theta,l_{ij}) \\
	& =\frac{p(\mathbf{x}_{ij}|z_{ijk}=1,\Theta) p(z_{ijk}=1|\Theta) p(z_{ijk}=1|l_{ij})}{\sum_{c=1}^K p(\mathbf{x}_{ij}|z_{ijc}=1,\Theta) p(z_{ijc}=1|\Theta) p(z_{ijc}=1|l_{ij})}\;,
	\end{split}
    \label{eq:train}
	\end{align}
	where $\Theta$ indicates the set of model parameters, $\{l_{ij}\}_{j=1,\ldots,N}$ are categorical manual labels assigned to image $i$ and $p(z_{ijk}=1|l_{ij})$ indicates the probability of voxel $j$ in image $i$ belonging to class $k$, given the manual label attributed to the same voxel. 
	
	A simple model for this, is 
	\begin{align}
	p(z_{ijk}=1|l_{ij}) = 
	\begin{cases}
	    \zeta_l,                      & \text{if } l_{ij}=k\\
	    \frac{1-\zeta_l}{K-1}, & \text{if } l_{ij}\neq k
	\end{cases}
	\end{align}
	where $\zeta_l$ is the sensitivity of the rater that generated the set of labels $\{l_{ij}\}_{j=1,\ldots,N}$ for image $i$. The problem of how to evaluate the performance of a manual or automated rater is not addressed here. For instance, a probabilistic scheme, which has been widely used to assess segmentation performance in medical imaging, is presented in \citet{warfield2004simultaneous}.
	
	\subsection{Combining diffeomorphic with affine registration}
	
	Anatomical shapes are very high dimensional objects. The diffeomorphic model described in the previous section can account for a significant amount of shape variability in the observed data. 	
	Nevertheless, it is still convenient, mainly for computational reasons, to combine such a local, high dimensional shape model with global, lower dimensional transformations, such as rigid body or affine transforms. In fact, by beginning to solve the registration problem from the coarsest deformation components (e.g. rigid body or affine), it is possible to ensure that the subsequent diffeomorphic registration starts from a good initial estimate of image alignment, that is to say closer to the desired global optimum \citep{lester1999survey}.

    This makes the optimisation problem faster to solve and at the same time it reduces significantly the chance of registration failure \citep{modersitzki2004numerical}. Indeed, it is relatively common for non-linear registration algorithms to perform poorly in the presence of a large translational or size mismatch between the reference and the target images  \citep{jenkinson2001global}.
	
	A possible parametrisation that combines affine and diffeomorphic transformations is 
	\begin{align}
	\bm{\xi}_i(\bm{y}) = \mathbf{T}_i \,\bm{\phi}_i(\bm{y}) + \mathbf{t}_i,\;~\forall \bm{y} \in \Omega_i\;,
	\label{eq:aff}
	\end{align}
	where $\bm{\xi}_i(\bm{y})$ is the resulting mapping from image of subject $i$ into the template space. Such a mapping is obtained by affine transforming the diffeomorphic deformation field $\bm{\phi}_i$. The transformation matrix $\mathbf{T}_i$ encodes nine degrees of freedom (rotation, zooming and shearing) and is computed via an exponential map  $\mathbf{T}_i = \exp (\mathbf{Q}_i(\mathbf{a}_i))$ with $\mathbf{Q}_i(\mathbf{a}_i) \in \mathfrak{ga}(3)$,  where $\mathfrak{ga}(3)$ is the Lie algebra for the affine group in three dimension $GA(3)$ and $\mathbf{a}_i$ is a vector of nine parameters \citep{ashburner2013symmetric}. Translations are modelled by the vector $\mathbf{t}_i \in \mathbb{R}^3$. The entire set of affine parameters is denoted as $\Theta_a=\{\mathbf{a}_i,\mathbf{t}_i\}_{i=1,\ldots,M}$.
	
	\subsection{Intensity model}
	\label{secIn}
	
	From a general probabilistic perspective, classification of tissue types based on MR signal intensities requires a model of the observed data that is capable of capturing the probability of occurrence of each signal sample value $\mathbf{x}_{ij}$, provided that the true labels are known. In other words, the problem breaks down into defining suitable conditional probabilities $p(\mathbf{x}_{ij}|z_{ijk}=	1)$, for each $k=\{1,\ldots,K\}$ and then applying Bayes rule to infer the posterior class probabilities. 
	
	In the model adopted here, image intensity distributions are represented as Gaussian mixtures, with the unknown mean $\bm{\mu}_{ik}$ and covariance matrix $\bm{\Sigma}_{ik}$ of each Gaussian component $k$, for subject $i$, being governed by Gaussian-Wishart priors \citep{bishop2006pattern,blaiotta2016variational}.
	
	Correction of intensity inhomogeneities is also performed within the same modelling framework and it involves multiplying the uncorrected intensities of each image volume by a bias field, which is modelled as the exponential of a weighted sum of discrete cosine transform basis functions \citep{styner2000parametric,unified}. Such an approach is conceptually equivalent to scaling the probability distributions of all Gaussian components by a local scale parameter, which is the bias itself, such that
	\begin{align}
	p(\mathbf{x}_{ij}|z_{ijk},\bm{\mu}_{ik},\bm{\Sigma}_{ik},\Theta_\beta) = \mathcal{N}(\mathbf{x}_{ij}|\bm{\hat{\mu}}_{ik},\bm{\hat{\Sigma}}_{ik})\;,
	\end{align}
	with
	\begin{align}
	\begin{split}
	\bm{\hat{\mu}}_{ik} &= \left(\text{diag}(\mathbf{b}_{ij})\right)^{-1}\bm{\mu}_{ik}\;,\\
	\bm{\hat{\Sigma}}_{ik} &= \left(\text{diag}(\mathbf{b}_{ij})\right)^{-1}\bm{\Sigma}_{ik} \left(\text{diag}(\mathbf{b}_{ij})\right)^{-1}\;,
	\end{split}
	\end{align}
	where $\Theta_\beta$ denotes the set of bias field parameters and $\mathbf{b}_{ij}$ is a $D$-dimensional vector representing the bias for subject $i$ at voxel $j$.
	
	\subsection{Graphical model}
	
\begin{figure}
	\centering
	\includegraphics[height=7cm]{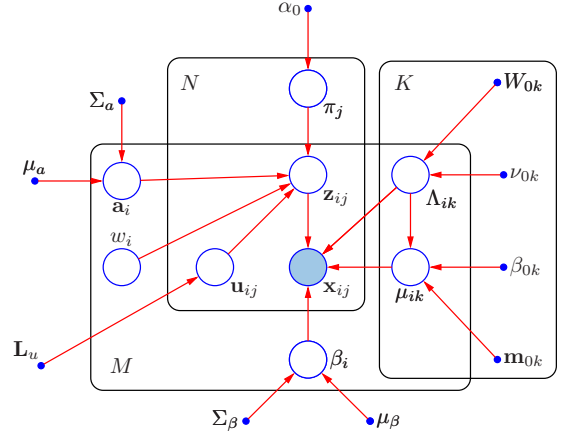}
	\caption{Graphical representation of the model adopted in this paper. Observed variables $\{\mathbf{x}_{ij}\}$ are represented by a filled circle. Latent variables $\{ \mathbf{z}_{ij}\}$ as well as model parameters are depicted as unfilled circles. Blue solid dots correspond to fixed hyperparameters. The so called plate notation is adopted to indicated repeated variables. Symbols referring to all variables and parameters are listed in table \ref{sym}.}
	\label{fig: graph2}
	\end{figure}

	A graphical representation of the model adopted in this paper is depicted in \Cref{fig: graph2}, while a legend of the symbols used to indicate the different variables can be found in table \ref{sym}.
	
	\begin{table*}
	\centering
	\begin{tabular}{c @{\hskip 0.28in} l}
	  \bf{Symbol} & \bf{Meaning}  \\
	   \midrule
	  $\mathbf{x}_{ij}$ & Observed image intensity at voxel $j$ for subject $i$.  \\
	  $\mathbf{z}_{ij}$ &  Vector of latent class membership probabilities. \\
	  $\bm{\pi}_j$ &   Tissue priors at voxel $j$.\\
	  $\bm{\mu}_{ik}$ & Mean intensity of class $k$ for subject $i$.   \\
	  $\bm{\Sigma}_{ik}$ & Covariance of intensities for class $k$ and subject $i$. \\
	  $\bm{W}_{0k}$ & Scale matrix of Wishart prior distribution on $\bm{\Lambda}_k = {(\bm{\Sigma}_k})^{-1}.$  \\
	  $\nu_{0k}$  &   Degrees of freedom of Wishart prior distribution on $\bm{\Lambda}_k$.  \\
	  $\bm{m}_{0k}$  & Mean of Gaussian prior distribution over $\bm{\mu}_k$  \\
	  $\beta_{0k}$  &    Scaling hyperparameter of Gaussian prior distribution over $\bm{\mu}_k$  \\
	  $\alpha_0$  &  Hyperparameter governing the Dirichlet prior on $\bm{\pi}$. \\
	  $\Theta_{\beta}$  &  Bias field parameters. \\
	  $\bm{\mu}_{\beta}$ & Prior mean of bias parameters.  \\
	  $\bm{\Sigma}_{\beta}$ & Prior covariance matrix of bias parameters. \\
	  $\Theta_{a}$  &  Affine transformation parameters. \\
	  $\bm{\mu}_{a}$ & Prior mean of affine transformation parameters.  \\
	  $\bm{\Sigma}_{a}$ & Prior covariance matrix of affine transformation parameters. \\
	  $\bm{w}_{i}$ &  Weights for rescaling the tissue priors. \\
	  $\mathbf{u}_{ij}$ &  Initial velocity at voxel $j$ for subject $i$. \\
	  $\bm{L}_{u}$ &  Differential operator to compute penalty on $\mathbf{u}_{i}$. \\
	  \bottomrule
	\end{tabular}
	\caption{}
	\label{sym}
	\end{table*}

	Given such a model, it is possible to define the following variational objective function $\mathcal{L}$, which constitutes a lower bound on the  logarithm of the marginal joint probability $p(\xbar{\mathbf{{X}}},\Theta_\beta,\Theta_a,\Theta_u,\Theta_\pi|\Theta_w)$, such that
	\begin{flalign}
	\log p(\xbar{\mathbf{{X}}},\Theta_\beta,\Theta_a,\Theta_u,\Theta_\pi|\Theta_w) \geq \mathcal{L}
	\end{flalign}
	and
	\begin{flalign}
	\begin{split}
	 \mathcal{L}= &\mathbb{E}_{\mathbf{Z},\Theta_{\mu},\Theta_{\Sigma}}[\log p(\xbar{\mathbf{{X}}}|\xbar{\mathbf{{Z}}},\Theta_{\mu},\Theta_{\Sigma},\Theta_{\beta})] \\ +&\mathbb{E}_{\mathbf{Z}}[\log p(\xbar{\mathbf{{Z}}}|\Theta_{\pi},\Theta_w,\Theta_{u},\Theta_a)] \\ +& \mathbb{E}_{\Theta_{\mu},\Theta_{\Sigma}}[\log p(\Theta_{\mu},\Theta_{\Sigma})] \\ +& \log p(\Theta_\pi) +\log p(\Theta_{\beta}) + \log p(\Theta_{a}) + \log p(\Theta_{u})
	\cr -&\mathbb{E}_{\mathbf{Z}}[\log q(\xbar{\mathbf{{Z}}})]  -\mathbb{E}_{\Theta_{\mu},\Theta_{\Sigma}}[\log q(\Theta_{\mu},\Theta_{\Sigma})]\;,
	\label{eq:LB1}
	\end{split}
	\end{flalign}
	where the expectations indicated as $\mathbb{E}_\mathbf{Z}$ and $\mathbb{E}_{\Theta_\mu,\Theta_\Sigma}$ are computed with respect to variational posterior distributions $q(\cdot)$ on the latent variables $\xbar{\mathbf{{Z}}}$ and on the Gaussian means and covariances $\{\Theta_\mu,\Theta_\Sigma\}$, respectively. Optimisation of $\mathcal{L}$, which provides optimal parameter and hyperparameter estimates, will be discussed in the following section.
	
	\subsection{Model fitting}
	
	The model described in the previous section can be fit to data sets of MR images by combining a variational expectation-maximisation (VBEM) algorithm with gradient based numerical optimisation techniques.
	
	Indeed, the VBEM algorithm described in \cite{blaiotta2016variational} is well-suited for addressing the model estimation problem discussed here, since it allows learning posterior distributions on the Gaussian mixture parameters, under the assumption that $q(\xbar{\mathbf{{Z}}},\Theta_\mu,\Theta_\Sigma)$ factorizes as $q(\xbar{\mathbf{{Z}}})q(\Theta_\mu,\Theta_\Sigma)$ \citep{bishop2006pattern}, and at the same time it is able to transfer the information encoded in such posteriors to estimate empirical intensity priors for each tissue type.
	
	Additionally, the algorithm proposed in this paper loops over all subjects in the population and, for each subject, it iterates over estimating the Gaussian posteriors, the bias field, the affine parameters and the initial velocities, which are all treated as conditional optimisations. Subsequently the tissue probability maps and intensity priors are updated and the whole cycle is repeated until convergence.
	
	Estimation of the bias field parameters $\Theta_\beta$ can be conveniently performed via non-linear optimisation techniques. Here the problem is solved using the Gauss-Newton method \citep{bertsekas1999nonlinear}, so as to maximise the objective function in \eqref{eq:LB1} with respect to $\Theta_\beta$. The resulting implementation is very similar to the one described in \citet{unified}, therefore further details are omitted here.
	Optimisation of the affine parameters $\Theta_a=\{\mathbf{a}_{i},\mathbf{t}_i\}_{i=1,\ldots,M}$ can also be carried out by means of a Gauss-Newton scheme and a brief description of the required computations can be found \ref{affine}. For the update of the weight parameters $\Theta_w$ we adopt the same strategy outlined in \citet{unified,blaiotta2016variational}.
	
	The following sections instead will present in detail the algorithmic scheme used to learn the average shaped tissue templates $\Theta_\pi=\{\bm{\pi}_{j}\}_{j=1,\ldots,N_\pi}$ and to estimate the set of initial velocity fields $\Theta_u=\{\mathbf{u}_{i}\}_{i=1,\ldots,M}$. 	
	
	\subsubsection{Updating the tissue priors}
	
	At each main iteration of the algorithm, the tissue priors $\Theta_\pi=\{\bm{\pi}_{j}\}_{j=1,\ldots,N_\pi}$ need to be updated, given the current estimates of all the other parameters, which are kept fixed for each individual in the population. 
	
	Considering only the terms in \eqref{eq:LB1} that depend on $\Theta_\pi$ gives the following objective function, which has to be maximised with respect to $\Theta_\pi$
	\begin{align}
	\begin{split}
	\mathcal{L}_\pi &= \mathbb{E}_{\mathbf{Z}}[\log p(\xbar{\mathbf{{Z}}}|\Theta_{\pi},\Theta_w,\Theta_{u},\Theta_a)] + \log p(\Theta_\pi) + \text{const}\\
	&= \sum_{i=1}^M \sum_{k=1}^K \bigintssss\limits_{\Omega_i}\gamma_{ik}(\bm{y})\log\left (\frac{w_{ik}\pi_{k} (\bm{\xi}_{i}(\bm{y}))}{\sum_{c=1}^K w_{ic}\,\pi_{c}(\bm{\xi}_{i}(\bm{y}))}\right) \text{d}\bm{y}\\ &+\log  p(\Theta_\pi) + \text{const}\;.
	\end{split}
	\label{ep:prnt}
	\end{align}
	
	It should be noted that the parameters $\Theta_\pi$ that need to be estimated are defined on the domain of the template $\Omega_\pi$, rather than on the individual spaces $\{\Omega_i\}_{i=1,\ldots,M}$. For this reason equation \eqref{ep:prnt}, which is a sum of integrals on the native domains, needs to be mapped to $\Omega_\pi$, by inverting the warps $\{{\bm{\xi}_i}\}_{i=1,\ldots,M}$, to give
	\begin{align}
	\begin{split}
	\mathcal{L}'_\pi =& \sum_{i=1}^M \sum_{k=1}^K \int\limits_{\Omega_\pi} \det\left(\frac{\partial \bm{\xi}_{i}^{-1}}{\partial\bm{y}}\right) \gamma_{ik}(\bm{\xi}_{i}^{-1}(\bm{y}))\log\left(\frac{w_{ik}\pi_{k} (\bm{y})}{\sum_{c}^K w_{ic}\,\pi_{c}(\bm{y})}\right) \text{d}\bm{y}\\ &+ \log p(\Theta_\pi) + \text{const}\;,
	\end{split}
	\label{eq:prupc}
	\end{align}
	where the determinants of the Jacobian matrices of the deformations are included to preserve volumes after the change of variables.
	
	Finally equation \eqref{eq:prupc} is discretised on a regular voxel grid, whose centres have coordinates $\{\mathbf{y}_j\}_{j=1,\ldots,N_\pi}$, to give
	\begin{align}
	\begin{split}
	\mathcal{L}'_\pi &= \sum_{i=1}^M \sum _{j=1}^{N_\pi} \sum_{k=1}^K \det(\mathbf{J}^{\bm{\xi}^{-1}}_{ij} )\, \gamma_{ik}(\bm{\xi}_{ij}^{-1})\,\log\left (\frac{w_{ik}\pi_{jk}}{\sum_{c=1}^K w_{ic}\,\pi_{jc}}\right) \\&+ \log p(\Theta_\pi) + \text{const}\;,
	\label{eq:prupc1}
	\end{split}
	\end{align}
	where 
	\begin{align}
    \bm{\xi}^{-1}_{ij}&=  \bm{\xi}_{i}^{-1}(\bm{y})\rvert_{\bm{y}=\mathbf{y}_j}\;,\\
	\det(\mathbf{J}^{\bm{\xi}^{-1}}_{ij}) &= \det\left(\left.\frac{\partial \bm{\xi}_{i}^{-1}(\bm{y})}{\partial\bm{y}}\right)\right\rvert_{\bm{y}=\mathbf{y}_j}\;,\\
    \pi_{jk} & =  \pi_k(\bm{y})\rvert_{\bm{y}=\mathbf{y}_j}\;.
	\end{align}
	
	The prior term $p(\Theta_\pi)$ is given by the following Dirichlet distribution 
	\begin{equation}
	p(\Theta_\pi)  = \prod_{j=1}^{N_\pi}\text{Dir}(\bm{\pi}_j|\bm{\alpha}_0) = C(\bm{\alpha}_0) \prod_{j=1}^{N_\pi} \prod_{k=1}^K \pi_{jk}^{\alpha_{0k}-1}\;.
	\end{equation}
	Maximising equation \eqref{eq:prupc1} is a constrained optimisation problem, subject to
	\begin{align}
	\sum_{k=1}^K \pi_{jk} =1\;, \, \forall j \in \{1,\ldots,N_\pi\}
	\label{eq:cons}
	\end{align}
	
	A closed form solution could be easily found if the rescaling weights $\{\bm{w}_i\}_{i=1,\ldots,M}$ were all equal to one. In such a case
	\begin{align}
	\begin{split}
	\mathcal{L}'_\pi =& \sum_{i=1}^M \sum _{j=1}^{N_\pi} \sum_{k=1}^K \det({\bm{J}}_{ij})\, \gamma_{ik}(\bm{\xi}_{ij}^{-1})\,\log\left (\pi_{jk}\right)\\& + \sum _{j=1}^{N_\pi} \sum_{k=1}^K  (\alpha_{0k}-1)\log p(\pi_{jk}) + \text{const}\;,
	\label{eq:prupcs}
	\end{split}
	\end{align}
	which could be maximised under the constraint \eqref{eq:cons}, by making use of Lagrange multipliers \citep{falk1967lagrange}, to give 
	\begin{align}
	\pi_{jk}= \frac{N_{jk} + \alpha_{0k} - 1}{\sum_{k=1}^K (N_{jk} + \alpha_{0k}) - K}\;,
	\end{align} 
	with ${N}_{jk} = \sum_{i=1}^M \det({\bm{J}}_{ij})\, \gamma_{ik}(\bm{\xi}_{ij}^{-1})$.

	This solution would provide maximum a posteriori point estimates of $\Theta_\pi=\{\bm{\pi}_{j}\}_{j=1,\ldots,N_\pi}$. However for this problem, it would also be possible to derive a full variational posterior distribution, which, like its prior, would take a Dirichlet form, with parameters $\bm{\alpha}_j = \bm{\alpha_0 } + \bm{N}_j$\;.
	
	When rescaling of the tissue priors is allowed the optimisation problem becomes more complex. The strategy adopted here consists in finding an approximate solution to the unconstrained optimisation problem by setting the derivatives of the objective function in \eqref{eq:prupc} to zero 
	\begin{align}
	\frac{\alpha_{0k}-1}{\pi_{jk}}  + \sum_{i=1}^M \det({\mathbf{J}}_{ij}^{\bm{\xi}^{-1}})\, \gamma_{ik}(\bm{\xi}_{ij}^{-1})\left( \frac{1}{\pi_{jk}} - \frac{w_{ik}}{\sum_{c=1}^K w_{ic}\pi_{jc}} \right) = 0\;.
	\end{align}
	
	Solving with respect to $\pi_{jk}$, under the simplifying assumption that the term $\sum_{c=1}^K w_{ic}\pi_{jc}$ can be treated as a constant, gives 
	\begin{align}
	\bar{\pi}_{jk} = \frac{N_{jk} + \alpha_{0k} - 1}{\sum_{i=1}^M\frac{ \det(\mathbf{J}^{\bm{\xi}^{-1}}_{ij})\, \gamma_{ik}(\bm{\phi}_{ij}^{-1})w_{ik}}{\sum_{c=1}^K w_{ic}\pi_{jc}}}\;.
	\end{align}

	Such a solution is then projected onto the constraining hyperplane, by preserving tissue proportions at each voxel
	\begin{align}
	\pi_{jk}=\frac{\bar{\pi}_{jk}}{\sum_{c=1}^K \bar{\pi}_{jc}}\;.
	\end{align} 

	Experimental testing of this strategy indicated that it gave a constant improvement of the objective function at a relatively cheap computational cost. Alternatively, iterative constrained non-linear optimisation techniques \citep{powell1978fast} could have been exploited to solve the template update problem.
	
	\subsubsection{Computing the deformation fields}
	
	Groupwise image alignment is achieved by optimisation of the variational objective function defined in \eqref{eq:LB1}, with respect to the parameters used to compute the deformations. This is equivalent to adopting the following image matching or similarity term 
	\begin{align}
	\begin{split}
	\mathcal{D} &= \mathbb{E}_{\mathbf{Z}}[\log p(\xbar{\mathbf{{Z}}}|\Theta_{\pi},\Theta_w,\Theta_{u},\Theta_a)] \\
	&= \sum_{i=1}^M \bigintsss\limits_{\bm{y}\in \Omega_i} \sum_{k=1}^K \gamma_{ik}(\bm{y})\log\left (\frac{w_{ik}\pi_{k} (\bm{\xi}_{i}(\bm{y}))}{\sum_{c=1}^K w_{ic}\,\pi_{c}(\bm{\xi}_{i}(\bm{y}))}\right) \text{d}\bm{y} \;.
	\end{split}
	\label{ep:prnt1}
	\end{align}
	Additionally, working on discretised image grids, with associated voxel centres $\{\mathbf{y}_{ij}\}_{j=1,\ldots,N_i}$, requires reformulating $\mathcal{D}$ as
	\begin{align}
	\begin{split}
	\mathcal{D} =\sum_{i=1}^M\sum_{j=1}^{N_i} \sum_{k=1}^K \gamma_{ijk}\log \frac{w_{ik} \pi '_{jk}}{{\sum_{c=1}^K w_{ic} \pi '_{jc}}}\;,
	\end{split}
	\end{align}
    with
    \begin{align}
    \pi'_{jk} &= \pi_k(\bm{\xi}_i(\bm{y}))\vert_{\bm{y}=\mathbf{y}_{ij}}\;. 
    \end{align}

	The penalty term for this groupwise image registration problem is given by 
	\begin{align}
	\begin{split}
	\mathcal{R} &=\mathcal{R}_{dif}+ \mathcal{R}_{af} = \log p(\Theta_u) + \log p(\Theta_a) \\& = - \frac{1}{2}\sum_{i=1}^{M}\left(||\mathbf{L_{u}}\mathbf{u}_i ||
	_{L^2}^2 + \mathbf{a}_i^T \bm{\Sigma_a}^{-1} \mathbf{a}_i\right) + \text{const}\;,
	\end{split}
	\end{align}
	with $\mathbf{u}_i$ being a $3 \times N_i$ dimensional vector of parameters used for representing the initial velocity field of image $i$ and $\mathbf{a}_i$ encoding affine deformation parameters used to compute the transformation in \eqref{eq:aff}. 
	
	For each image $i$ in the data set, updating the corresponding initial velocity field, given the current estimates of the templates and all the other model parameters, involves optimising the following objective function
	\begin{align}
	\begin{split}
	\mathcal{E}_{dif}^{(i)} = &\mathcal{D}^{(i)} +  \mathcal{R}_{dif}^{(i)}  \\=& \sum_{j=1}^{N_i} \sum_{k=1}^K \gamma_{ijk}\log \frac{w_{ik} \pi_{k} (\bm{\xi}_{ij})}{{\sum_{c=1}^K w_{ic} \pi_{c} (\bm{\xi}_{ij})}} - \frac{1}{2}||\mathbf{L_{u}}\mathbf{u}_i ||
	_{L^2}^2\;,
	\end{split}
	\end{align}
	with respect to $\mathbf{u}_i$, under the following deformation model 
	\begin{align}
	\bm{\xi}_{ij} = \bm{\xi}_i(\mathbf{y}_{ij}) = \mathbf{T}_i \,\bm{\phi}_i(\mathbf{y}_{ij}) + \mathbf{t}_i\;,
	\end{align}
	where $\bm{\phi}_i$ is a diffeomorphism computed via geodesic shooting \citep{ashburner2011diffeomorphic} from the corresponding initial velocity field $\mathbf{u}_i$.
	
	Here image registration is solved via Gauss-Newton optimisation, which requires computing both the first and second derivatives of the objective function \citep{hernandez2008gauss}. Such derivatives can be found in \ref{diffeo}. This leads to a very high dimensional inverse problem, which unfortunately cannot be solved via numerical matrix inversion, since this would be prohibitively expensive from a computational point of view. The approach adopted in this work consists in treating this optimisation as a partial differential equation problem, which can efficiently be solved using multigrid methods \citep{modersitzki2004numerical}. In particular, we adopt the same full multigrid implementation as in \citet{ashburner2007fast}.
	
	\section{Validation and Discussion}
	
	In this section we will present results obtained by applying the presented modelling framework to real brain and cervical cord MR scans acquired with different imaging protocols, as well as to synthetic MR head volumes. Both qualitative and quantitative measures will be provided to assess the behaviour of the proposed approach.
	
	\subsection{Template construction}
	
	As discussed in \Cref{methods}, the proposed method can serve to learn prior tissue probability maps from cross-sectional imaging data sets. In this paper we mainly explore the performance of such a framework to address the quest for integrated brain and spinal cord neuromorphometric tools, even if, given the generality of the presented approach, many more applications should in principle be possible.
	
	\subsubsection{Data}
	\label{data}

	The input data for training the model was obtained from three different databases, two of which are freely accessible for download, thus ensuring that the results presented here could readily be compared to those produced by competing algorithms for medical image registration or segmentation.
	
	\paragraph{OASIS data set}
	
	The first data set consists of thirty five T1-weighted MR scans from the OASIS (Open Access Series of Imaging Studies) database \citep{marcus2007open}. The data is freely available from the web site \url{http://www.oasis-brains.org}, where details on the population demographics and acquisition protocols are also reported. Additionally, the selected thirty five subjects are the same ones that were used within the 2012 MICCAI Multi-Atlas Labeling Challenge \citep{landman2012miccai}. 
	
	\paragraph{Balgrist data set}
	
	The second data set consists of brain and cervical cord scans of twenty healthy adults, acquired at University Hospital Balgrist with a 3T scanner (Siemens Magnetom Verio). Magnetisation-prepared rapid acquisition gradient echo (MPRAGE) sequences, at 1 $mm$ isotropic resolution, were used to obtain T1-weighted data, while PD-weighted images of the same subjects were acquired with a multi-echo 3D fast low-angle shot (FLASH) sequence, within a whole-brain multi-parameter mapping protocol \citep{weiskopf2013quantitative,helms2008quantitative}.

	\paragraph{IXI data set}
	
	The third and last data set comprises twenty five T1-, T2- and PD-weighted scans of healthy adults from the freely available IXI brain database, which were acquired at Guy's Hospital, in London, on a 1.5T system (Philips Medical Systems Gyroscan Intera). Additional information regarding the demographics of the population, as well as the acquisition protocols, can be found at \url{http://brain-development.org/ixi-dataset}.
	
	\vspace{5mm}
	
	The complete data set therefore consists of eighty multispectral scans of healthy adults, obtained with fairly diverse acquisition protocols and using scanning systems produced by different vendors. 
	
	Unfortunately, not all the three modalities of interest (T1-, T2- and PD-weighted) are available for all of the subjects. 
	To circumvent the difficulties arising from the presence of missing imaging modalities, without neglecting any of the available data (indeed deletion of entries with missing data is still, in spite of its crudity, a common statistical practice), the Gaussian mixture modelling approach discussed in \cite{blaiotta2016variational} was generalised by introducing an additional variational posterior distribution over the missing data points. 

	In practice, the resulting variational EM scheme iterates over first estimating an approximated posterior distribution on the unknown image intensities, secondly updating the sufficient statistics of the complete (observed and missing) data and finally computing variational posteriors on the Gaussian mixture parameters. Additional computational details relative to this strategy are provided in \ref{miss_data}.
	
	In synthesis, it was possible to fit the generative groupwise model described in this paper to the entire data set, in spite of having different imaging modalities available from the different acquisition sites. This is indeed a very common scenario in real life medical imaging problems, therefore it should be actively addressed by processing or modelling solutions that claim to be applicable to large population data \citep{van2015does}.
	
    Manual brain labels are freely available for all images in data set one. Such labels have been generated and made public by Neuromorphometrics, Inc. (\url{http://Neuromorphometrics.com}) under academic subscription and they provide a fine parcellation of cortical and non cortical structures, for a total of 139 labels across the brain.
	
	Part of this label data was used for training of the model while the remainder was left out for testing and validation. In particular, brain labels of twenty out of the thirty five OASIS subjects were used to create gray and white matter ground truth segmentations, which were provided as training input for semisupervised model fitting. 

	Similarly, spinal cord manual labels were created for forty subjects (twenty from data set two and twenty from data set three). Such labels were randomly split in half for training and half for subsequent test analyses.
	Due to the limited resolution of the data it was not possible to manually delineate gray and white matter within the spinal cord. For this reason, each voxel classified as spinal cord in the training data was allowed to be assigned either to the gray or to the white matter tissue classes, based on the fit of its intensity value to the underlying Gaussian mixture model, as outlined in equation \eqref{eq:train}.
	
	Analogously, in spite of having defined only one gray matter training label, two distinct gray matter classes were introduced in the mixture model (top two rows in \Cref{fig: TPMs}), to best capture the corresponding distribution of image intensities, which is poorly represented by a single Gaussian component, as opposed to the distribution of white matter intensities. Also in this case, membership probabilities of the labelled training data were computed based on the corresponding intensity values, by making use of equation \eqref{eq:train}.

	\subsubsection{Tissue templates and intensity priors}
	
\begin{figure*}
	\centering
	\subfloat{\includegraphics[trim={1cm 0 0 0},clip,height=20cm]{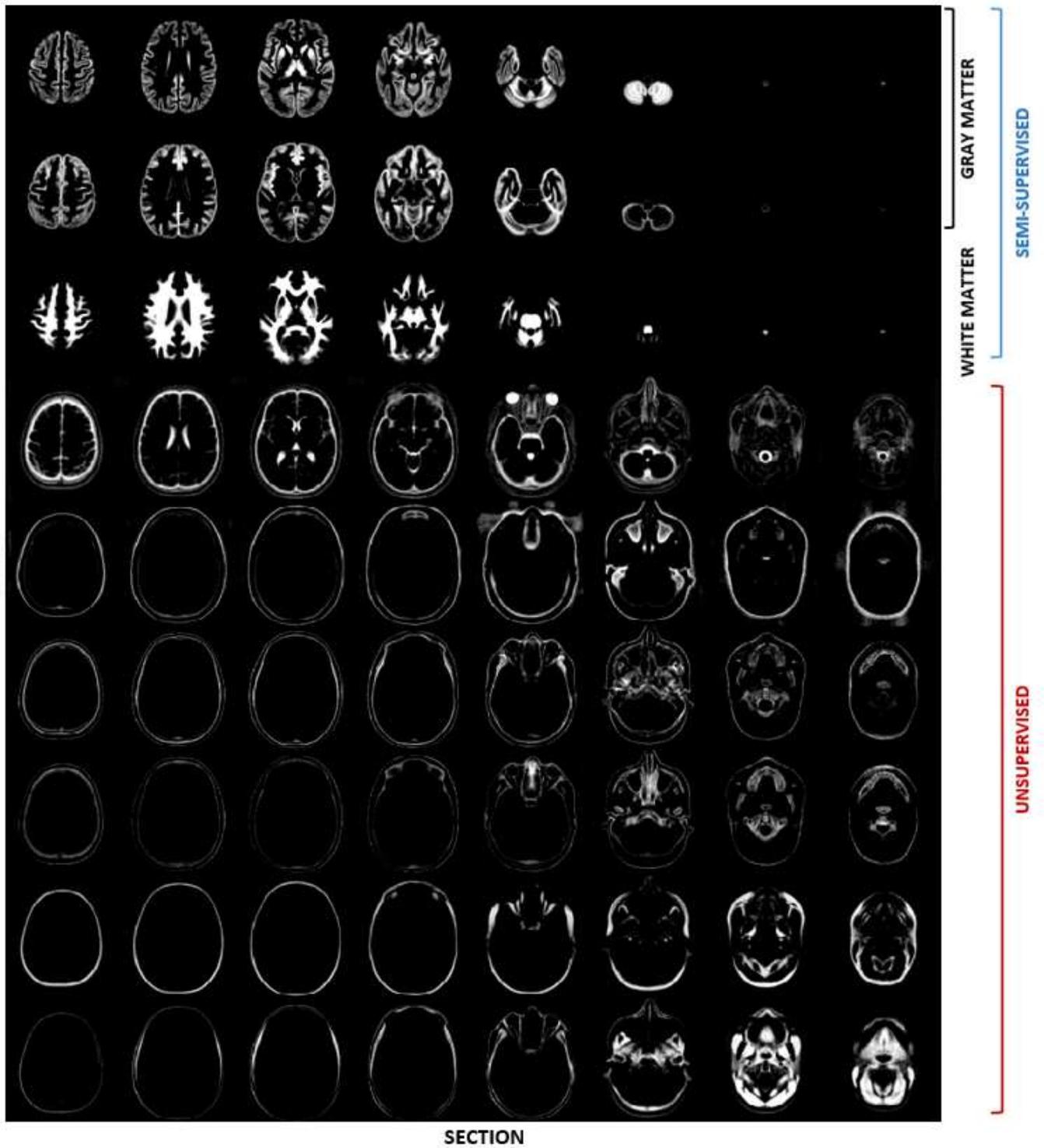}}
	\caption[]{Tissue probability maps obtained by applying the presented groupwise generative model to a multispectral data set comprising brain and cervical cord scans of eighty healthy adults, from three different databases.}
	\label{fig: TPMs}
	\end{figure*}

	The tissue probability maps obtained by applying the modelling framework presented in this paper to the data set described above are depicted in \Cref{fig: TPMs}. The total number of tissue classes used for this experiment is equal to twelve but three classes, representing air in the background, are not shown. 
	
	In particular, \Cref{fig: TPMs} shows how one of the two gray matter classes (first row) best fits the subcortical nuclei and also includes voxels affected by partial volume effects at the interface between gray and white matter, while the second one (second row) is more representative of cortical structures, with the presence of partial volume effects generated by the juxtaposition of gray matter and CSF. The third row in \Cref{fig: TPMs} shows the white matter class, which also includes most of the brainstem and the spinal cord.
	
	The remaining tissue classes were estimated in a purely unsupervised way. Therefore a non ambiguous anatomical interpretation is not straightforward. 
	
	Tissue class four (fourth row) mainly contains CSF, even if other tissues are also present, especially in the neck area. This should be attributed to the lack of CSF training labels as well as to a poor multivariate coverage of the cervical region in the available data. In fact, data from the OASIS set is truncated around the first cervical vertebra.  The T1-weighted scans of the IXI data set cover up to the C2/C3 vertebral level, but the corresponding T2- and PD-weighted scans do not extend beyond the brainstem. Indeed, only the data from the second database (Balgrist hospital) provides more than one modality covering up to around the fourth cervical vertebra. In this case though, additional difficulties arose from poor inter-modality alignment of the data, a problem that turned out to be particularly severe in the cervical region and that, given its non-linearity, could not be fully compensated for by affine inter-modality coregistration. 
	
	Bone tissue is also not easily identifiable from the data available for this experiment, but it could have potentially been much better extracted by incorporating some CT scans into the training data.
	
	Fat and soft tissues are mainly represented in the last two classes (bottom two rows in \Cref{fig: TPMs}).
	
	\begin{figure}
	\centering
	\subfloat[Gray matter]{\label{gray_tpm}\includegraphics[height=5.4cm]{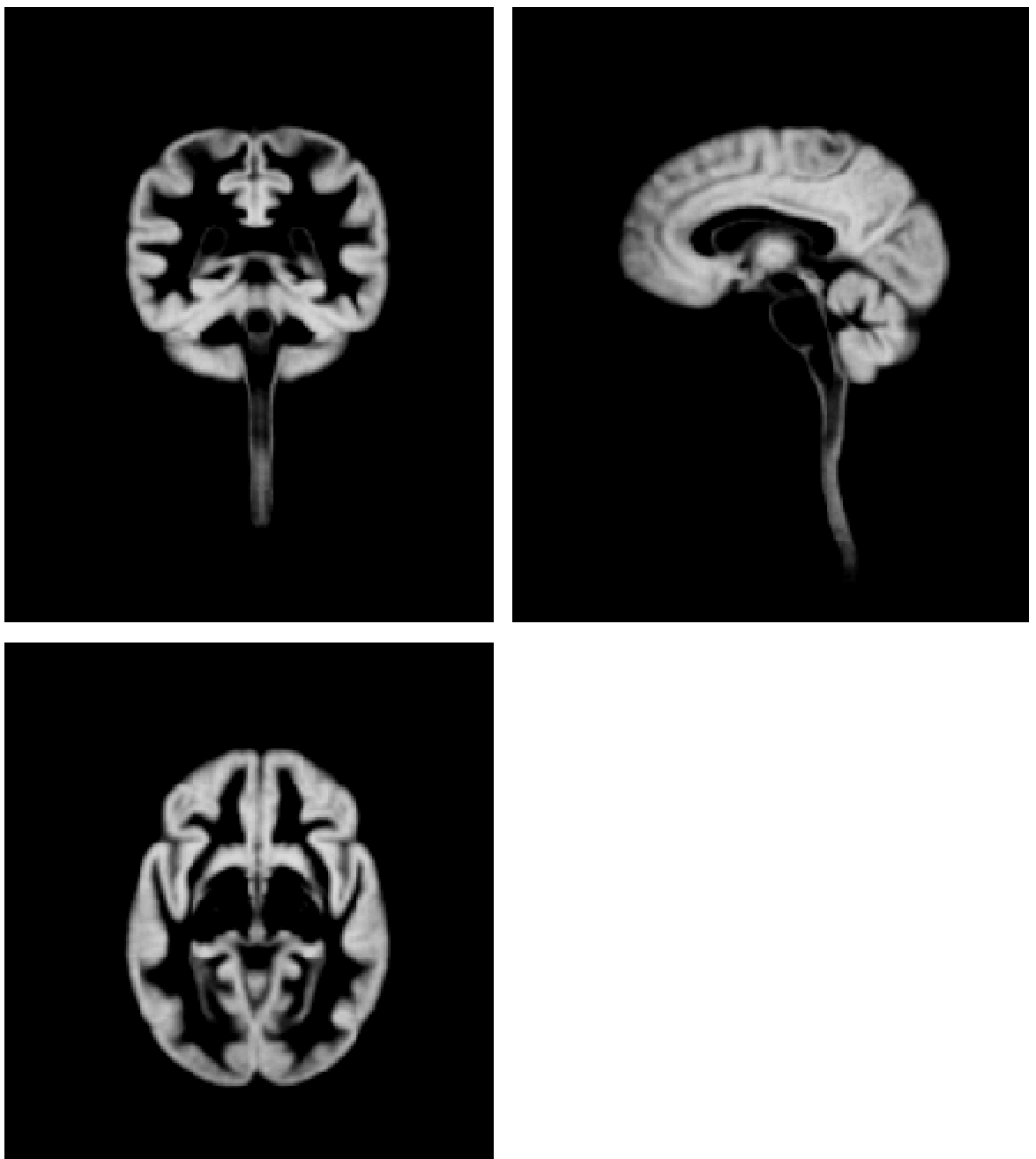}}\\
	\subfloat[White matter]{\label{white_tpm}\includegraphics[height=5.4cm]{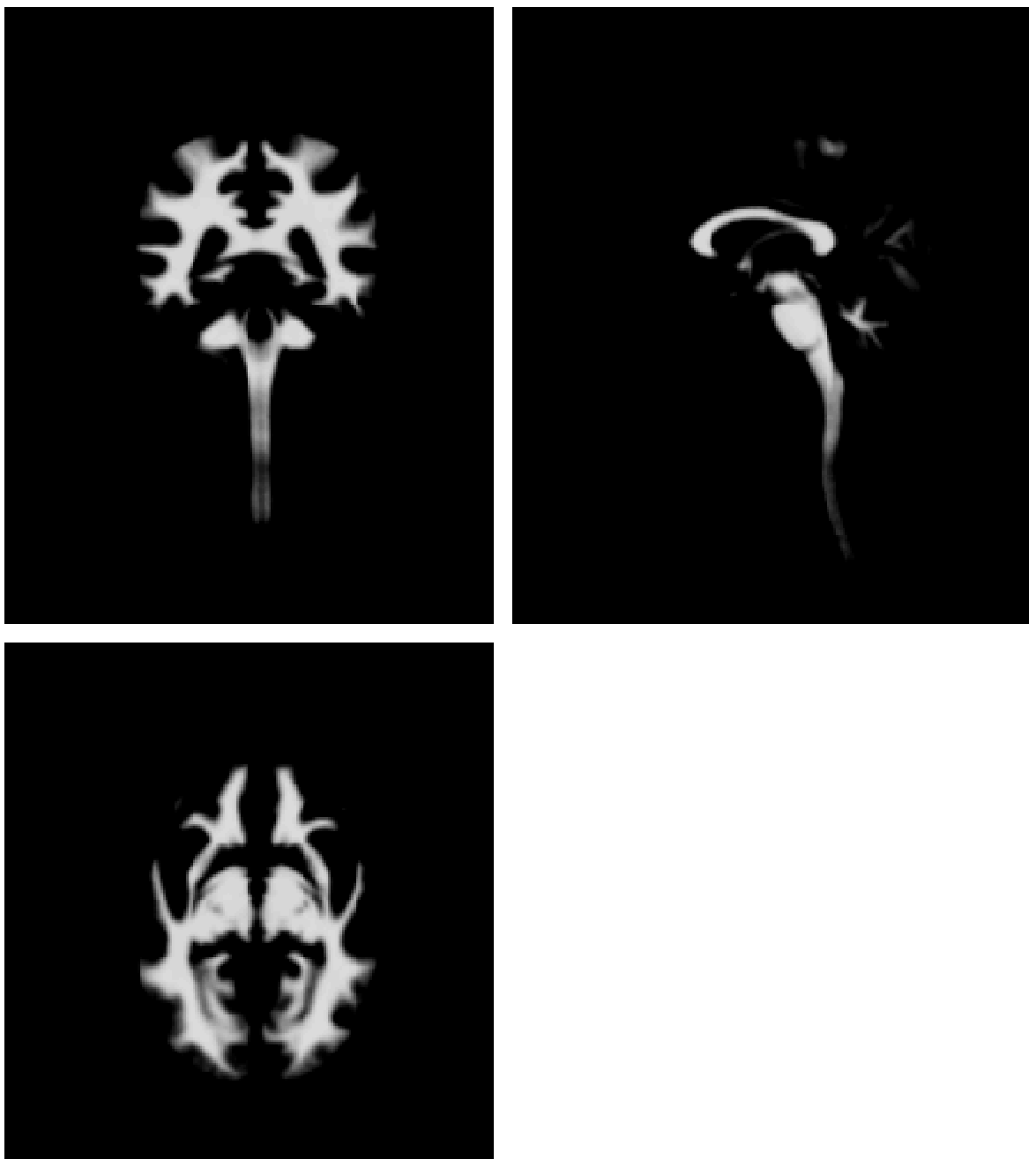}}
	\caption[]{Brain and spinal cord tissue probability maps of gray \subref{gray_tpm} and white \subref{white_tpm} matter, constructed as described in this paper.}
	\label{fig: gwtpm}
	\end{figure}
	
	\Cref{fig: gwtpm} illustrates orthogonal views of the gray and white matter tissue probability maps, where the gray matter map is obtained by evaluating the sum of the two top classes in \Cref{fig: TPMs}. 
	
\begin{figure}
	\centering
    \label{gray}\includegraphics[width=7cm]{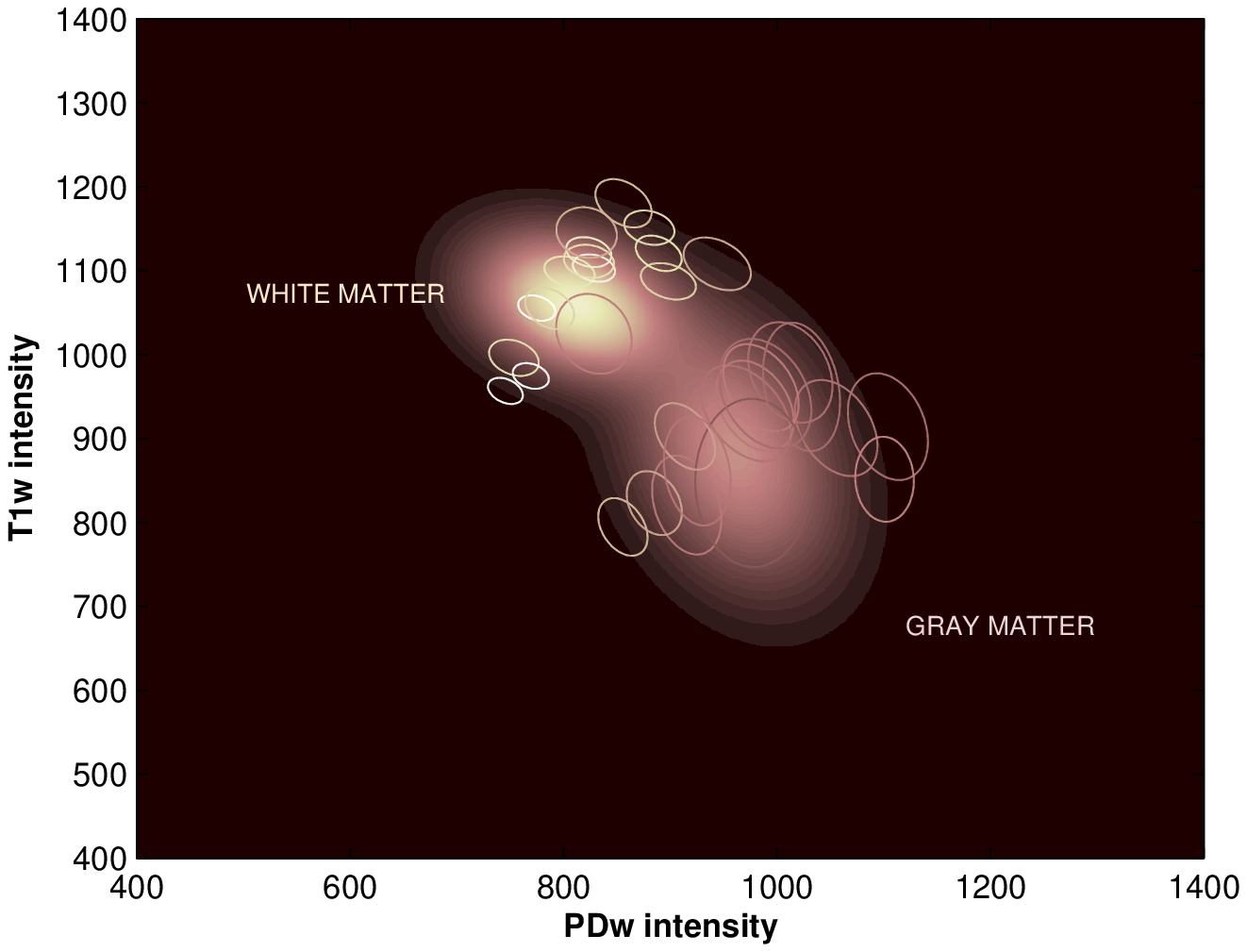}
	\caption[]{Prior distribution over the mean intensity of gray and white matter, in T1- and PD-weighted data.}
	\label{fig: intensity_priors}
	\end{figure}

	The empirical Bayes learning procedure, introduced in \citet{blaiotta2016variational} to estimate suitable prior distributions for the parameters of the Gaussian mixture model, was applied here to the same data used to construct the templates. Some of the results are summarised in figure \ref{fig: intensity_priors}, where the estimated empirical prior distributions on the mean intensity of gray and white matter are depicted, with overlaid contour plots showing some of the individual posteriors (randomly selected across the entire population).
	
	Such results indicate that the proposed empirical Bayes learning scheme can serve to capture, not only the variability of mean tissue intensity across subjects for each of the modalities of interest, but also the amount of covariance between such modalities. Information of this sort can potentially be used in a number of different frameworks, for solving problems such as tissue segmentation, pathology detection or image synthesis.
	
	\subsubsection{Validity of groupwise registration}
	
	The performance of groupwise registration achieved by the presented algorithm was assessed by computing pairwise overlap measures for all possible couples of spatially normalised test images (i.e. images whose ground truth labels were not used for training the model). The Dice score coefficient was chosen as a metric of similarity.
	
\begin{figure*}
	\centering
	\caption[Accuracy of groupwise registration achieved by the presented method, compared to the performance of ANTs]{Accuracy of groupwise registration achieved by the presented method, compared to the performance of ANTs, for different neural regions. Stars indicate statistically significant differences between the two methods, assessed by means of paired t-tests without correcting for multiple comparisons.}
	\subfloat{\includegraphics[trim={3.2cm 2cm 1.3cm 0cm},clip,width=0.45\textwidth,height=0.95\textheight]{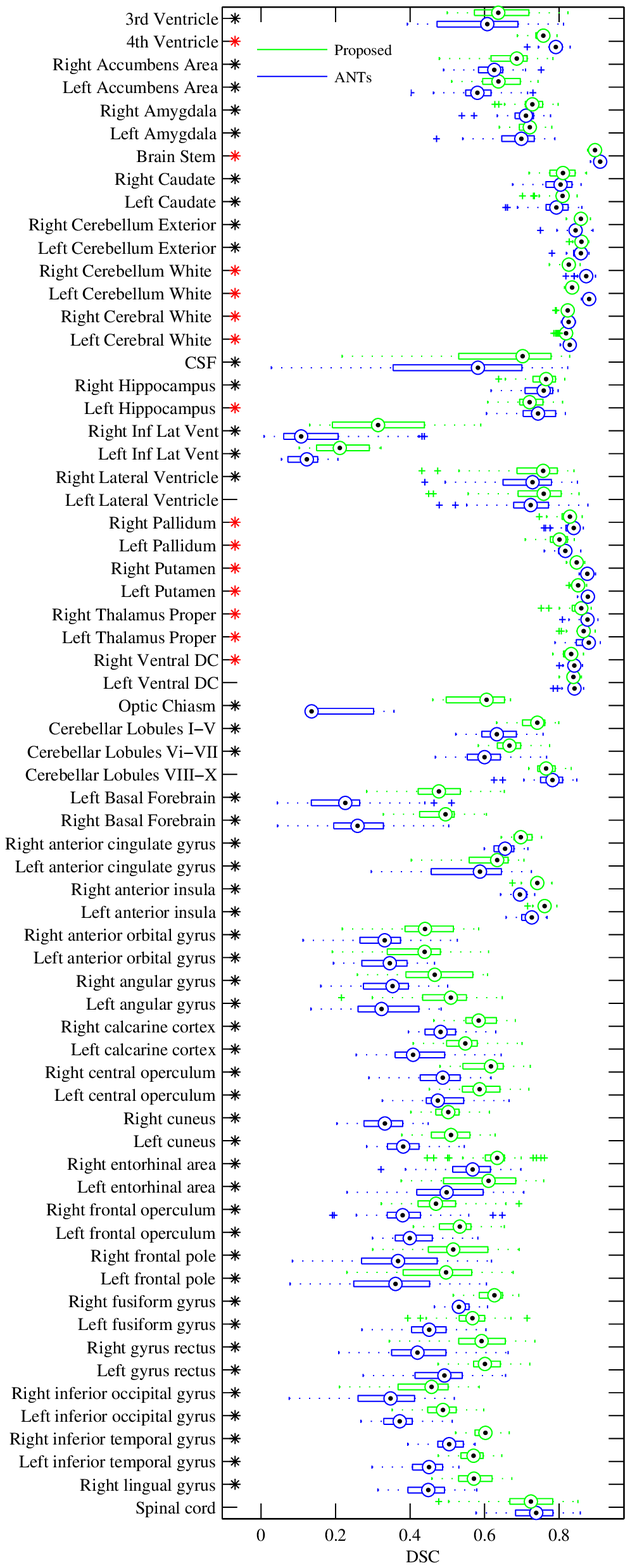}}
	\subfloat{\includegraphics[trim={3.2cm 2cm 1.3cm 0cm},clip,height=0.95\textheight,,width=0.5\textwidth]{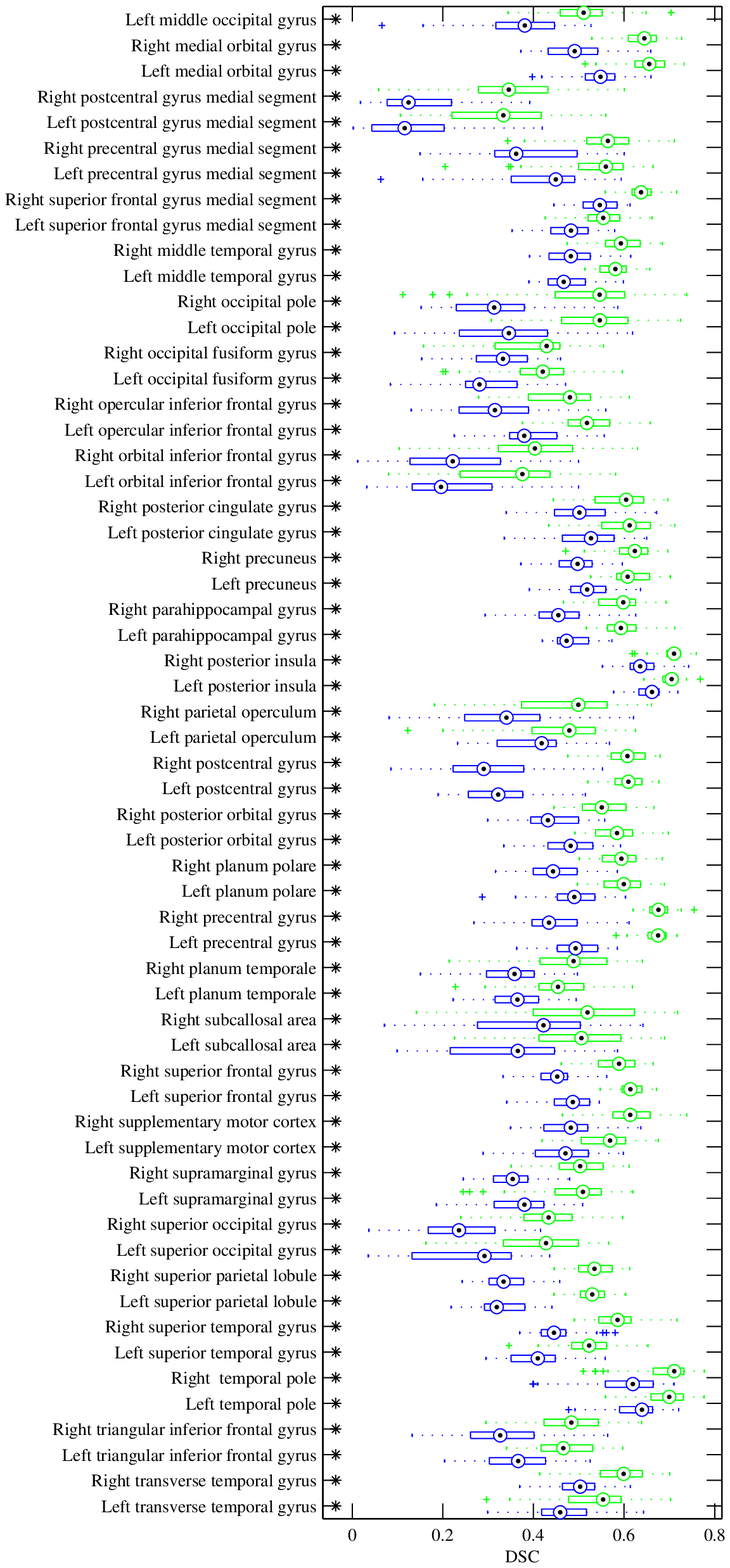}}
	\label{fig: regac1}
	\end{figure*}

	Results are summarised in figure \ref{fig: regac1}, where the accuracy of the algorithm presented here is compared to that achieved by the method described in \cite{avants2010optimal}, whose implementation is publicly available, as part of the Advanced normalisation Tools (ANTs) package, through the web site \url{http://stnava.github.io/ANTs/}. Indeed, the symmetric diffeomorphic registration framework implemented in ANTs has established itself as the state-of-the-art of medical image nonlinear spatial normalisation \citep{klein2009evaluation}.
	
	A number of options can be customised within the template construction framework distributed with ANTs. The experiments, whose results are reported here, were performed with the settings recommended in the package documentation for brain MR data, which are also reported in table \ref{tab: ANTS}.
	
	\begin{table}
	\setlength{\tabcolsep}{2pt}
	\centering
	\caption[Options selected to perform groupwise registration with ANTs]{Options selected to perform groupwise registration with ANTs, using the \texttt{antsMultivariateTemplateConstruction} script provided with the ANTS package.}
	\begin{tabular}{lc}
	\toprule
	\bf{Option} &\bf{Value}\\
	\midrule
	Similarity Metric &  \bf{Cross-correlation (CC)} \\ 
	Transformation model &  \bf{Greedy SyN (GR)} \\ 
	Initial rigid body &  \bf{yes} \\ 
	N4 Bias Correction &  \bf{yes} \\ 
	Number of resolution levels &  \bf{4} \\ 
	Number of iterations &  $\mathbf{100 \times 70 \times 50 \times 10}$ \\ 
	Gradient step &  \bf{0.2} \\ 
	Number of template updates &  \bf{4} \\ 
	\bottomrule 
	\label{tab: ANTS}
	\end{tabular} 
	\end{table}
	
	Results of this validation analyses indicate that the method presented here, in spite of not being as accurate as ANTs for aligning some subcortical brain structures (e.g. thalamus, putamen, pallidum and brainstem), provided significantly better overlap when registering cortical regions, as assessed by means of paired t-tests with a significance threshold of 0.05 and without correcting for multiple comparisons. No statistically significant differences were found between the two methods with respect to registration of the spinal cord. 
	
	\subsubsection{Accuracy of tissue classification}
	
	The accuracy of tissue classification achieved by the method presented in this paper was first evaluated on test data that was used to create the templates but without providing manual labels for training the model during atlas construction. The aim in this case was to determine to which extent the proposed method can capture relevant features of the training data, when manual labels are not provided, by learning from few annotated examples.
	Dice scores \footnote{The Dice score over two sets $A$ and $B$ is defined as $DSC = 2  \frac{|A \cap B|}{|A| + |B|}$\;.} were computed to compare the automated segmentations produced via semisupervised groupwise model fitting, with the ground truth, obtained by merging all the gray and white matter brain structures (labels) into two tissue classes respectively, and by considering the spinal cord as a third separate class. 
	
	\begin{figure}
	\centering
	\includegraphics[width=8cm]{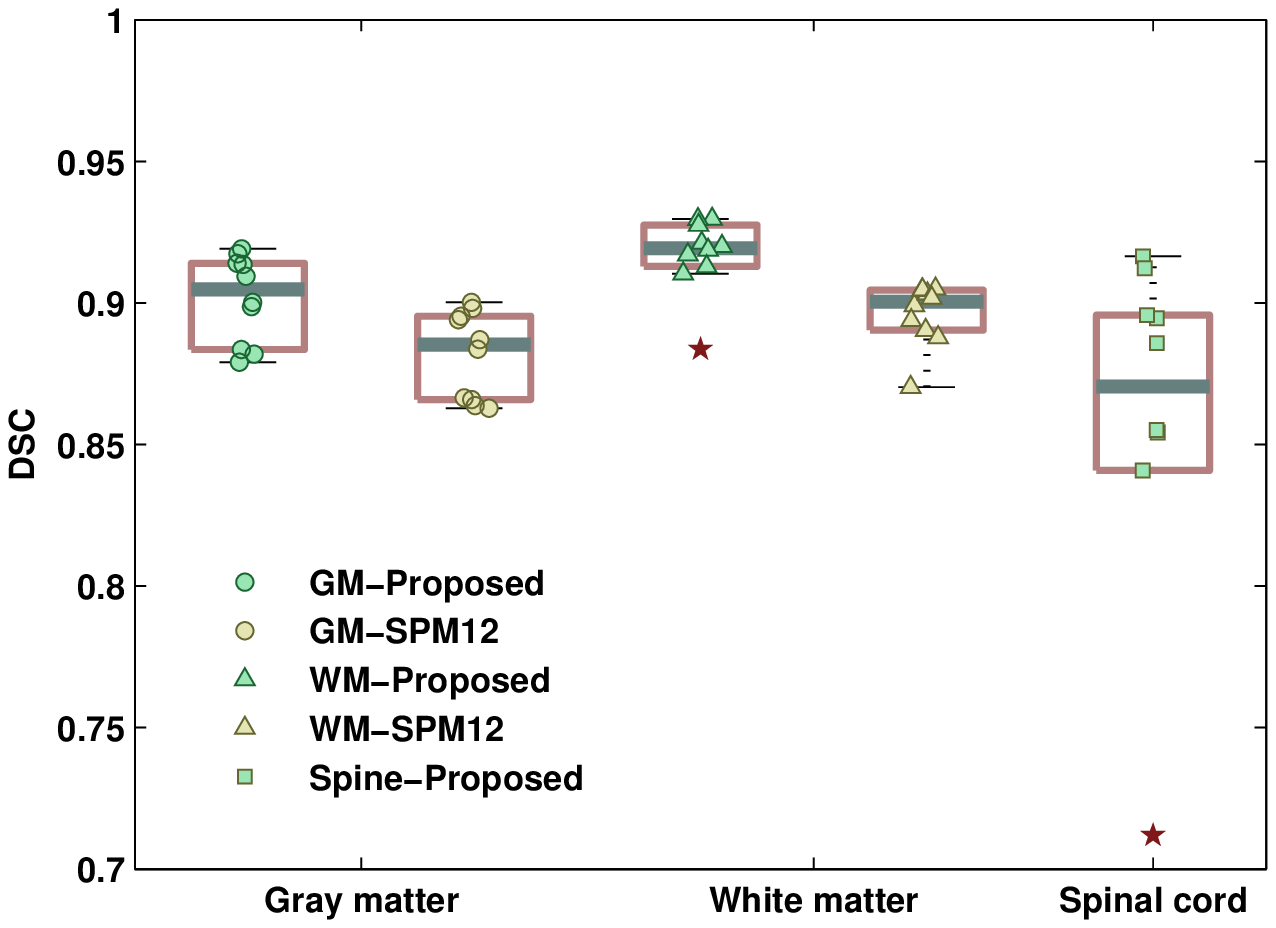}
	\caption[Brain and spinal cord segmentation accuracy of the presented method]{Brain and spinal cord segmentation accuracy of the presented method.}
	\label{Spine_seg}
	\end{figure}
	
	The probabilistic gray and white matter segmentations of the brain were thresholded at 0.5, in order to obtain binary label maps, directly comparable to the ground truth. To derive binary cord segmentations instead, the sum of gray and white matter posterior belonging probabilities was first computed in a subvolume containing the neck only, and then thresholded at 0.5. 
	
	Results are summarised in \Cref{Spine_seg}, which shows the distributions of Dice scores obtained for brain gray matter, brain white matter and spinal cord.
	 
	Such results were then compared to those produced by the brain segmentation algorithm implemented in SPM12, using the standard tissue probability maps distributed with the SPM software. Results of these analyses, which are summarised in \Cref{Spine_seg}, indicate that the population specific atlases constructed with the method presented here enable higher tissue classification accuracy, at least for test data drawn from the same population that the model was trained on but whose labels were not exploited for training. A potential source of bias in the results of this experiment is the fact that the test data was actually employed for constructing the atlases, even if the corresponding labels were not seen by the algorithm. However a more cautious k-fold cross-validation, which would have required constructing multiple templates, was not practical in this case due to the expensive computational cost of groupwise model fitting. 
	
	Such results however seem to suggest that the model presented in this paper could potentially be useful to create templates with the purpose of capturing the peculiar anatomical features of those populations that are poorly represented by standard anatomical atlases \citep{tang2010construction,fillmore2015age}, such as young or elderly populations, diseased populations, or individuals belonging to different ethnic groups. 
	
	This would not only lead to more accurate segmentation results, but as a direct consequence, also increase the reliability of subsequent data analyses, which build models of the segmented data to infer or predict clinically meaningful information.	
	
	\subsection{Modelling unseen data}
	
	Further validation experiments were performed to quantify the accuracy of the framework described in this paper 
	to model unseen data, that is to say data that was not included in the atlas generation process.
	
	Such experiments were performed on synthetic T1-weighted brain MR scans from the Brainweb database (\url{http://brainweb.bic.mni.mcgill.ca/}), generated using a healthy anatomical model. 
	
	\subsubsection{Accuracy of bias correction}
	
	A healthy adult brain MR model was processed by means of the algorithm discussed here, using the head and neck templates previously constructed as tissue priors. Different noise and bias field levels were added to the uncorrupted synthetic data, to test the behaviour of the proposed modelling scheme in different noise (1\%, 3\%, 7\%) and bias conditions (20\% and 40\%). 
	
	The noise in these simulated images has Rayleigh statistics in the background and Rician statistics in the signal regions and its level is computed as a percent standard deviation ratio, relative to the MR signal, for a reference tissue \citep{cocosco1997brainweb}.
	
	Regarding the bias field instead, 20\% bias is modelled as a smooth field in the range [0.9, 1.1] while 40\% bias is obtained by rescaling of the 20\% field, so as to range between 0.8 and 1.2\,. 
	
	\begin{table}
	\setlength{\tabcolsep}{12pt}
	\centering
	\caption[]{Pearson's correlation coefficients between the ground truth bias fields and those estimated by the presented algorithm, for simulated T1-weighted data.}
	\begin{tabular}{cc|ccc}
	\toprule
	&& \multicolumn{3}{c}{\bf{Noise}} \\
	&& \bf{1\%} & \bf{3\%}& \bf{7\%} \\ 
	\midrule
	&\bf{20\%} & 0.86 & 0.86 & 0.70\\ 
	\multirow{-2}{*}{\bf{Bias}} & \bf{40\%} &  0.72 & 0.72 & 0.51\\ 
	\bottomrule 
	\label{tab: Bias}
	\end{tabular} 
	\end{table}
	
	\Cref{tab: Bias} reports the Pearson product-moment correlation coefficients between the ground truth and the estimated bias fields, for the different bias ranges and noise levels.
	Results indicate that the similarity between the estimated and true bias decreases for more intense non-uniformity fields and higher noise levels. 
	
	Indeed this is not surprising, as the penalty term, which enforces smoothness of the bias field, has a greater impact in determining the shape of the estimated bias when the non-uniformity fields have a larger dynamic range. Nevertheless, results reported in the following section will show how this increased mismatch between the estimated and true bias, for higher non-uniformities, does not seem to affect the accuracy of tissue segmentation. On the other hand, the accuracy of bias correction is directly related to the amount of noise corrupting the data, mainly due to how this affects the precision associated with estimation of the Gaussian mixture parameters. For a comparison of these results with the performance of SPM12 bias correction on simulated T1-weighted scans from the Brainweb database see \cite{blaiotta2016variational}.
	
	\subsubsection{Accuracy of tissue classification}
	
	For the same data the accuracy of tissue classification was also evaluated, by comparing the similarity between the estimated gray and white matter segmentations and the underlying anatomical  model. 
	
\begin{figure}
	\centering
    \label{whites}\includegraphics[trim={0.cm 0cm 0cm 0},width=8cm]{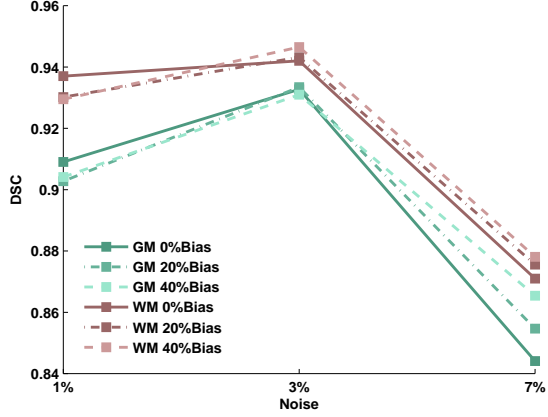}
	\caption[Dice scores between estimated and ground truth brain segmentations]{Dice scores between the estimated and ground truth segmentations for brain white matter and brain gray matter, under different noise and bias conditions, for synthetic T1-weighted data.}
	\label{DSC_seg}
	\end{figure}

	Results are reported in \Cref{DSC_seg}, which shows the Dice score coefficients obtained under different bias and noise conditions.
	
	The Brainweb database has been extensively used in the neuroimaging community to validate MR image processing algorithms. Therefore the results reported here should be directly comparable to the performance of many brain segmentation techniques present in the literature.
	
	\section{Conclusions}

	This paper presented a comprehensive generative framework for modelling cross-sectional MR data sets, which is intended to enable simultaneous morphometric analyses of brain and cervical spinal cord data. 

	From a theoretical perspective, such a framework relies on variational probability density estimation techniques to model the observed data (i.e. MR signal intensities). Additionally, a hierarchical modelling perspective is proposed, where observations from a population of subjects are used to construct empirical intensity priors, which can then serve to inform models of new data.

	Shape modelling is performed via groupwise diffeomorphic registration, thus ensuring bijective (i.e. one-to-one) differentiable mappings between anatomical configurations \citep{miller2004computational}. Such an approach enables a rigorous mathematical encoding of anatomical shapes via deformable template matching \citep{christensen1996deformable}, therefore providing a quantitative framework for the analysis of shape variation and covariation.
	
	Data for training the method was collected from three different databases, two of which are publicly accessible to the research community. Results of validation experiments performed both on training and unseen test data indicate that the presented framework is suitable to perform integrated brain and cervical cord computational morphometrics. 

	Thus, the proposed algorithm represents a concrete solution to extract volumetric and morphometric information from large structural neuroimaging data sets, in a fully automated manner. At the same time it provides outputs that could be readily interpreted, for instance via statistical hypothesis testing, with the ultimate goal of comparing different populations, treatment effects etc. \citep{ashburner2000voxel}. 

	\appendix
	
		\section{Derivatives of the lower bound with respect to the affine parameters}
	\label{affine}
	
	The affine parameters, for each subject $i$, can be estimated (i.e. optimised) in a Gauss-Newton fashion, so as to maximise of the following objective function
	\begin{align}
	\begin{split}
	\mathcal{E}_{af}^{(i)} &=  \mathcal{D}^{(i)} + \mathcal{R}_{af}^{(i)}  \\ &= \sum_{j=1}^{N_i} \sum_{k=1}^K \gamma_{ijk}\log \frac{w_{ik} \pi_{k} (\bm{\xi}_{ij})}{{\sum_{c=1}^K w_{ic} \pi_{c} (\bm{\xi}_{ij})}} - \frac{1}{2}\mathbf{a}_i^T \bm{\Sigma_a}^{-1} \mathbf{a}_i \;,
	\end{split}
	\end{align}
	with respect to $\mathbf{a}_i$.	
	
	The gradients and Hessians, which are useful to solve this problem are reported below. 
	In particular, for the matching term, the following derivatives need to be computed
	
	\begin{align}
	\frac{\partial\mathcal{D}^{(i)}}{\partial\mathbf{a}_{i}}  =  &\sum_{j=1}^{N_i}\sum_{k=1}^{K}  \left(\gamma_{ijk} - \frac{w_{ik}\, \pi_{k} (\bm{\xi}_{ij})}{{\sum_{c=1}^K w_{ic}\, \pi_{c} (\bm{\xi}_{ij})}}\right)\,\bm{g}_{jk}^\pi\;,
	\end{align}
	where $\bm{g}_{jk}^\pi$ is defined as
	\begin{align}
	\bm{g}_{jk}^\pi=\mathbf{B}_i^T \left([\bm{\phi}_{ij}\,,1] \otimes \nabla \left[\log \left(\pi_{k} (\bm{\xi}_{ij})\right)\right]\right)\;,
	\end{align}
	with
	\begin{align}
	\mathbf{B}_i^T = \frac{\partial\mathbf{S}_i}{\partial \mathbf{a}_i}\;,
	\end{align}
	and
	\begin{align}
	\mathbf{S}_i = 
	\begin{bmatrix}
	\mathbf{T}_i & \mathbf{t}_i\\
	0                   & 1
	\end{bmatrix}
	\;.
	\end{align}
	
	\begin{align}
	\begin{split}
	\frac{\partial^2\mathcal{D}^{(i)}}{\partial\mathbf{a}_{i}^2}  = & \sum_{j=1}^{N_i}\left(\sum_{k=1}^{K} \frac{w_{ik}\, \pi_{k} (\bm{\xi}_{ij})}{{\sum_{c=1}^K w_{ic}\, \pi_{c} (\bm{\xi}_{ij})}}\,\bm{g}_{jk}^\pi\right) \\& \times \left(\sum_{k=1}^{K} \frac{w_{ik}\left(\pi_{k} (\bm{\xi}_{ij})\right)}{{\sum_{c=1}^K w_{ic}\left(\pi_{k} (\bm{\xi}_{ij}\right))}}\,\bm{g}_{jk}^\pi \right)^T \\
	-& \sum_{j=1}^{N_i}\sum_{k=1}^{K} \frac{w_{ik}\, \pi_{k} (\bm{\xi}_{ij})}{{\sum_{c=1}^K w_{ic}\, \pi_{c} (\bm{\xi}_{ij})}}\,\bm{g}_{jk}^\pi \left(\bm{g}_{jk}^\pi\right)^T\;.
	\end{split}
	\end{align}
	
	Gradients and Hessians of the penalty term are instead given by	
	\begin{align}
	\frac{\partial\mathcal{R}_{af}^{(i)}}{\partial\mathbf{a}_{i}}  =-\bm{\Sigma_a}^{-1} \mathbf{a}_i  \;,
	\end{align}
	
	\begin{align}
	\frac{\partial^2\mathcal{R}_{af}^{(i)}}{\partial\mathbf{a}_{i}^2}  = -\bm{\Sigma_a}^{-1}\;.
	\end{align}

	\section{Derivatives of the lower bound with respect to the initial velocities}
	\label{diffeo}
	
	Optimisation of the initial velocities, for each image $i$, requires maximising the following objective function
	\begin{align}
	\begin{split}
	\mathcal{E}_{dif}^{(i)} &=  \mathcal{D}^{(i)} + \mathcal{R}_{dif}^{(i)}  \\ &= \sum_{j=1}^{N_i} \sum_{k=1}^K \gamma_{ijk}\log \frac{w_{ik} \pi_{k} (\bm{\xi}_{ij})}{{\sum_{c=1}^K w_{ic} \pi_{c} (\bm{\xi}_{ij})}} - \frac{1}{2}\sum_{i=1}^{M}||\mathbf{L_{u}}\mathbf{u}_i ||
	_{L^2}^2  \;,
	\end{split}
	\label{eq:Er}
	\end{align}
	with respect to $\mathbf{u}_i$.
	
    Here, we report the first and second derivatives of this objective function, which are useful to solve the registration problem using gradient-based techniques, such as the Gauss-Newton algorithm.
    
    The gradient of the matching term $\mathcal{D}^{(i)}$ with respect to $\mathbf{u}_{i}$ is given by
	\begin{align}
	\begin{split}
	\frac{\partial \mathcal{D}^{(i)}}{\partial \mathbf{u}_{i}} =& \sum_{k=1}^{K} \gamma_{ijk} \frac{\partial}{\partial \mathbf{u}_{i}} \bigg(\log \frac {w_{ik}\, \pi_{k} (\bm{\xi}_{i})}{\sum_{c=1}^K w_{ic}\, \pi_{c} (\bm{\xi}_{i})}\bigg)\\
	=&  \sum_{k=1}^{K} \gamma_{ijk}\,  \left(\bm{g}_k^\pi-
	 \sum_{c=1}^K \frac{w_{ic}\, \pi_{c} (\bm{\xi}_{i})}{\sum_{c=1}^K w_{ic}\, \pi_{c} (\bm{\xi}_{i})} \bm{g}_c^\pi\right)\;,
	\end{split}
	\end{align}
	which, making use of $\sum_{k=1}^K \gamma_{ijk} =1\,$, can be rewritten as 
	\begin{align}
	\frac{\partial \mathcal{D}^{(i)}}{\partial \mathbf{u}_{i}}= &\sum_{k=1}^{K}  \left(\bm{\gamma}_{ik} - \frac{w_{ik}\, \pi_{k} (\bm{\xi}_{i})}{{\sum_{c=1}^K w_{ic}\, \pi_{c} (\bm{\xi}_{i})}}\right)\,\bm{g}_k^\pi\;,
	\end{align}
	where $\bm{g}_k^\pi$ is computed, at each voxel $j$, by
	\begin{align}
	\bm{g}_{jk}^\pi=\left(\mathbf{T}_i,\mathbf{J}^{\bm{\xi}}_{ij}\right)^T \nabla \left[\log \left(\pi_{k} (\bm{\xi}_{ij})\right)\right]\;,
	\end{align}
	and $\mathbf{J}^{\bm{\xi}}_{i}$ indicates the Jacobian matrix of $\bm{\xi}_{ij}$.
	
	An approximated positive semidefinite Hessian of $\mathcal{D}$ can instead be computed by discarding the second derivatives of the logarithm of tissue priors 
	\begin{equation}
	\frac{\partial^2}{{\partial \bm{y}}^2}\log \left(\frac{w_{ik}\left(\pi_{k} (\bm{\xi}_{i}(\bm{y}))\right)}{{\sum_{c=1}^K w_{ic} \left(\pi_{c}(\bm{\xi}_{i}(\bm{y}))\right)}}\right) = 0\;,\forall \bm{y} \in \Omega_i\;,
	\end{equation}
	to give
	\begin{align}
	\begin{split}
	\frac{\partial^2 \mathcal{D}^{(i)}}{{\partial\mathbf{u}_{i}}^2} =& \left(\sum_{k=1}^{K} \frac{w_{ik}\, \pi_{k} (\bm{\xi}_{i})}{{\sum_{c=1}^K w_{ic}\, \pi_{c} (\bm{\xi}_{i})}}\,\bm{g}_k^\pi\right) \\ & \times \left(\sum_{k=1}^{K} \frac{w_{ik}\left(\pi_{k} (\bm{\xi}_i)\right)}{{\sum_{c=1}^K w_{ic}\left(\pi_{k} (\bm{\xi}_i\right))}}\,\bm{g}_k^\pi \right)^T \\
	& - \sum_{k=1}^{K} \frac{w_{ik}\, \pi_{k} (\bm{\xi}_{i})}{{\sum_{c=1}^K w_{ic}\, \pi_{c} (\bm{\xi}_{i})}}\,\bm{g}_k^\pi \left(\bm{g}_k^\pi\right)^T\;.
	\end{split}
	\end{align}
	
	Finally, the first and second derivatives of the penalty term $\mathcal{R}$, which are also required to optimise \eqref{eq:Er}, can be computed by
	\begin{equation}
	\frac{\partial\mathcal{R}^{(i)}_{dif}}{\partial\mathbf{u}_{i}} = -\mathbf{L_u}^\dagger\mathbf{L_u}\mathbf{u}_{i}\;,
	\end{equation}

	\begin{equation}
	\frac{\partial^2\mathcal{R}^{(i)}_{dif}}{{\partial\mathbf{u}_{i}}^2} = -\mathbf{L_u}^\dagger\mathbf{L_u}\;.
	\end{equation}
	
	\section{Variational Gaussian mixtures: inference of missing data}
	\label{miss_data}
	
	The variational Bayes EM algorithm for fitting Gaussian mixture models, described in \cite{blaiotta2016variational}, can be generalised to handle the case where some components of the $D$-dimensional observation $\mathbf{x}_j$ are missing. 
	
	Having denoted 
	\begin{equation}
	\mathbf{x}_j=
	\begin{bmatrix}
	\mathbf{o}_j   \\    \mathbf{h}_j
	\end{bmatrix}\;,
	\end{equation}
	with $\mathbf{o}_j$ being the observed data and $\mathbf{h}_j$ the missing data, the Gaussian likelihood $p(\mathbf{x}_j|z_{jk}=1,\bm{\mu}_k,\bm{\Sigma}_k)$ can be expressed as 
	\begin{align}
	p(\mathbf{x}_j|z_{jk}=1,\bm{\mu}_k,\bm{\Lambda}_k)
	=\mathcal{N}\left(\begin{bmatrix}
	 \mathbf{o}_j   \\    \mathbf{h}_j
	\end{bmatrix}\Bigg\rvert
	\begin{bmatrix}
	\bm{\mu}_k^o   \\    \bm{\mu}_k^h
    \end{bmatrix}\;,
	\begin{bmatrix}
	\bm{\Lambda}_k^{o,o}   & \bm{\Lambda}_k^{o,h} \\  \bm{\Lambda}_k^{o,h}   & \bm{\Lambda}_k^{h,h} 
	\end{bmatrix}\right)\;,
	\end{align}
	by making use of block matrix notation to partition the mean vector $\bm{\mu}_k$ and the precision matrix $\bm{\Lambda}_k$.	

	In this case $\mathbf{h}_j$ is treated as an unobserved random variable. Thus, in a variational Bayes setting, an additional posterior factor can be introduced for each missing data point $\mathbf{h}_j$ to give
	\begin{align}
	\begin{split}
	q(\mathbf{H},\mathbf{Z},{\Theta}_{\mu},{\Theta}_{\Sigma}) =  &q(\mathbf{H})q(\mathbf{Z})q({\Theta}_{\mu},{\Theta}_{\Sigma})\\
	 = &q(\mathbf{Z}) q({\Theta}_{\mu},{\Theta}_{\Sigma})\prod_{j=1}^N q(\mathbf{h}_j)\;.
	\end{split}
	\end{align}

    Making use of the general result $q_{\hat{s}}(\Theta_{\hat{s}})\propto\exp(\mathbb{E}_ {s\neq \hat{s}} [\log p(\mathbf{X},\Theta)])$ \citep{bishop2006pattern}, an approximated posterior on the missing data point $\mathbf{h}_j$ can be computed by
    \begin{align}
    \begin{split}
    \log q(\mathbf{h}_j)  =&  \mathbb{E}_{\mathbf{Z},\Theta_\mu,\Theta_\Sigma}\left[\log p(\mathbf{x}_j,\mathbf{z}_j,\Theta_\mu,\Theta_\Sigma|\Theta_\pi)\right] + \text{const}  \\
     =& \mathbb{E}_{\mathbf{Z},\Theta_\mu,\Theta_\Sigma}\left[\log p(\mathbf{x}_j|\mathbf{z}_j,\Theta_\mu,\Theta_\Sigma)\right] +\mathbb{E}_{\mathbf{Z}}\left[\log p(\mathbf{z}_j|\Theta_\pi)\right] \\ 
     &  + \mathbb{E}_{\Theta_\mu,\Theta_\Sigma}\left[\log p(\Theta_\mu,\Theta_\Sigma)\right] + \text{const}\;,
    \end{split}
    \label{eq:qh}
    \end{align}
    where $\Theta_\pi$ denotes the mixing proportion parameter set, treated here via maximum likelihood, and $p(\Theta_\mu,\Theta_\Sigma)$ is a conjugate Gaussian-Wishart prior on the means and covariances of the model. 
    
    Ignoring the terms independent from $\mathbf{h}_j$, equation \eqref{eq:qh} can be rewritten as
    \begin{align}
    	\begin{split}
    	\log q(\mathbf{h}_j)  =& \sum_{k=1}^K \gamma_{jk}\,\mathbb{E}_{\Theta_\mu,\Theta_\Sigma} \left[\log \mathcal{N}(\mathbf{x}_{j}|\boldsymbol{\mu}_{k},\boldsymbol{\Sigma}_{k})\right ] + \text{const} \\
     =&   \frac{1}{2}\sum_{k=1}^K \gamma_{jk}\mathbf{h}_j^T \mathbb{E}_{\Theta_\mu,\Theta_\Sigma}\left[\bm{\Lambda}_k^{h,h}\right] \mathbf{h}_j\\
    & + \sum_{k=1}^K \gamma_{jk}\mathbf{h}_j^T\mathbb{E}_{\Theta_\mu,\Theta_\Sigma}\left[\bm{\Lambda}_k^{o,h}\right]\left( \mathbf{o}_j - \mathbb{E}_{\Theta_\mu,\Theta_\Sigma}\left[ \bm{\mu}_k^o\right]\right)\\
    &-  \sum_{k=1}^K \gamma_{jk}\mathbf{h}_j^T\mathbb{E}_{\Theta_\mu,\Theta_\Sigma}\left[\bm{\Lambda}_k^{h,h}\right]\mathbb{E}_{\Theta_\mu,\Theta_\Sigma}\left[ \bm{\mu}_k^h\right]+\text{const}\;.
    	\end{split}
    	\end{align}
    
    The previous equation indicates that the unobserved value $\mathbf{h}_j$ is drawn from a Gaussian mixture distribution with mixing proportions equal to the posterior (after having observed $\mathbf{o}_j$) membership probabilities $\{\gamma_{jk}\}_{k=1,\ldots,K}$, while the Gaussian means $\{\mathbf{n}_{jk}\}_{k=1,\ldots,K}$ and covariances $\{\mathbf{P}_{jk}\}_{k=1,\ldots,K}$ are given by
    \begin{align}
    \label{eq:n}
    \begin{split}
    \mathbf{n}_{jk} =& \mathbb{E}_{\Theta_\mu,\Theta_\Sigma}\left[ \bm{\mu}_k^h\right] + \left(\mathbb{E}_{\Theta_\mu,\Theta_\Sigma}\left[\bm{\Lambda}_k^{h,h}\right]\right)^{-1} \\&\times \mathbb{E}_{\Theta_\mu,\Theta_\Sigma}\left[\bm{\Lambda}_k^{o,h}\right](\mathbb{E}_{\Theta_\mu,\Theta_\Sigma}\left[\bm{\mu}_k^{o}\right] - \mathbf{o}_j)\;,
    \end{split}
    \end{align}
    \begin{align}
    \mathbf{P}_{k} = \mathbb{E}_{\Theta_\mu,\Theta_\Sigma}\left[\bm{\Lambda}_k^{h,h}\right]\;.
    \label{eq:P}
    \end{align}
    
    Given the posteriors $q(\mathbf{Z})$ and $q(\mathbf{H})$, the following sufficient statistics of $\mathbf{X}$ can be computed
    \begin{align}
    \bm{s}_{1k} &= 
    		\begin{bmatrix}
    	 		\sum_{j=1}^N\gamma_{jk}\mathbf{o}_j   \\   \sum_{j=1}^N\gamma_{jk} \mathbf{n}_{jk}
    		\end{bmatrix}\;,\\
    \bm{S}_{2k} &= 
    		\begin{bmatrix}
                 \sum_{j=1}^N\gamma_{jk}\mathbf{o}_j \mathbf{o}_j^T & \sum_{j=1}^N\gamma_{jk}\mathbf{o}_j \mathbf{n}_{jk}^T\\
    		\sum_{j=1}^N\gamma_{jk}\mathbf{n}_{jk} \mathbf{o}_{j}^T & \sum_{j=1}^N\gamma_{jk}\left(\mathbf{n}_{k} \mathbf{n}_{jk}^T + (\mathbf{P}_{k})^{-1}\right) 
    		\end{bmatrix}\;.
    \end{align}
    
    Once such sufficient statistics have been evluated, they can be used to update the Gaussian-Wishart posteriors $q({\Theta}_{\mu},{\Theta}_{\Sigma})$ in the exact same way as in \cite{blaiotta2016variational}. Such posteriors are in turn used to compute the expectations that appear in equations \eqref{eq:n} and \eqref{eq:P}, in an iterative EM fashion.
	
	\section*{Acknowledgments}
	Claudia Blaiotta is co-funded by UCL and Zurich Balgrist Hospital, as part of the UCL 'Impact' award scheme. This research was supported by Wings for Life - Spinal Cord Research Foundation.
	The Wellcome Trust Centre for Neuroimaging is supported by core funding from the Wellcome Trust (091593/Z/10/Z).
	The OASIS project was funded by the National Institutes of Health grants P50 AG05681, P01 AG03991, R01 AG021910, P50 MH071616, U24 RR021382, R01 MH56584. The IXI project was supported by the EPSRC grant GR/S21533/02.
	
	\bibliographystyle{plainnat} 
	\section*{References}
    \bibliography{Biblio}

\end{document}